\documentclass[10pt,twocolumn,letterpaper]{article}

\usepackage[pagenumbers]{iccv} % To force page numbers, e.g. for an arXiv version

\definecolor{iccvblue}{rgb}{0.21,0.49,0.74}
\usepackage[pagebackref,breaklinks,colorlinks,allcolors=iccvblue]{hyperref}
\usepackage{multirow}
\usepackage{soul}
\usepackage{minted}

\usepackage{listings}
\usepackage{xcolor} % for color customization

\def\paperID{5579} % *** Enter the Paper ID here
\def\confName{ICCV}
\def\confYear{2025}

\definecolor{lightblue}{rgb}{0.85, 0.95, 1.0}

\newcommand{\boldred}[1]{\textcolor{red}{\textbf{#1}}}

\title{Disconnect to Connect: A Data Augmentation Method for Improving Topology Accuracy in Image Segmentation}

\begin{document}
	\author{Juan Miguel Valverde$^{1,2,3}$ \quad Maja Østergaard$^3$ \quad Adrian Rodriguez-Palomo$^3$ \\ Peter A. S. Vibe$^3$ \quad  Nina K. Wittig$^3$ \quad Henrik Birkedal$^3$ \quad Anders Bjorholm Dahl$^1$ 
        \vspace{0.3cm}        \\
		$^1$DTU Compute, Technical University of Denmark, Denmark\\
        $^2$A.I. Virtanen Institute, University of Eastern Finland, Finland\\
        $^3$Department of Chemistry and iNANO, Aarhus University, Denmark \\
        {\tt\small  \{jmvma,abda\}@dtu.dk \quad \{majaoester,adrian.rodriguez,petervibe\}@inano.au.dk \quad hbirkedal@chem.au.dk } \\
        }
\maketitle
\begin{abstract}
Accurate segmentation of thin, tubular structures (e.g., blood vessels) is challenging for deep neural networks. These networks classify individual pixels, and even minor misclassifications can break the thin connections within these structures.
Existing methods that focus on improving topology accuracy, such as topology loss functions, require very precise training labels with the correct topology, which is difficult to obtain.
Low image resolution and contrast can further complicate the annotation by causing these structures to appear disconnected.
We present CoLeTra, a data augmentation method that improves topology accuracy, even on datasets with topologically inaccurate training labels.
CoLeTra achieves this by artificially breaking the structures in the images while maintaining the original labels, incorporating to the models the prior knowledge that structures can be continuous even if they appear disconnected.
Our extensive experiments, involving different architectures, loss functions, and datasets, demonstrate that CoLeTra leads to segmentations topologically more accurate while often improving the Dice coefficient and clDice.
We also release a dataset specifically suited for image segmentation methods with a focus on topology accuracy.
CoLetra's code can be found at \url{https://jmlipman.github.io/CoLeTra}.
\end{abstract}   
\section{Introduction}
Despite the remarkable potential of deep learning for image segmentation \cite{minaee2021image}, accurate segmentation of thin, tubular structures, such as axons, airways, and blood vessels, remains a significant challenge.
This is partly due to the inherent pixel-wise classification nature of standard deep learning models, where small misclassifications that hardly increase the loss, can lead to segmentations with discontinuities, impairing the quantification \cite{hu2022learning}.
For instance, connected-component analysis of structures with broken connections would misinterpret each fragment as a distinct structure, resulting in overestimating their number.
An incorrect number of connected components would then compromise other measurements, such as average length, density, and directionality.

\begin{figure}
  \centering
  \includegraphics[width=0.5\textwidth]{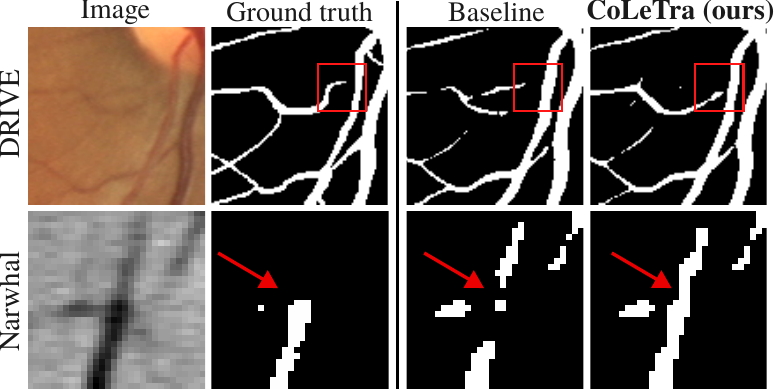}
  \caption{Left: Annotations missing important areas, compromising the structure's true topology. Right: CoLeTra improves topology accuracy.} \label{fig:teaser}
\end{figure}

An important line of research has focused on emphasizing structures' connectivity with topology loss functions \cite{hu2019topology,clough2020topological,hu2022learning,liao2023segmentation,shit2021cldice}.
These are loss functions that aim to produce segmentations that match the number of connected components, holes, and cavities of the ground truth, prioritizing topology accuracy over slight improvements in non-topology metrics, such as Dice coefficient.
However, these topology loss functions require topologically-precise labels, an often unrealistic requirement in areas such as medical imaging and material science \cite{shi2024survey}.
Due to the low image resolution and contrast, it is challenging to annotate the images because the tubular structures may appear disconnected (see \Cref{fig:teaser}).
In these cases with topologically-inaccurate labels, the loss functions inadvertently reinforce the absence of a structure that we know should exist based on prior knowledge about connectivity.
Furthermore, most topology loss functions rely on persistence homology \cite{oner2023persistent,stucki2023topologically,xu2024semi,zhang2025toposculpt}, soft-skeletonization \cite{shit2021cldice,viti2022coronary}, or distance transforms \cite{shi2024centerline,liu2024enhancing}, which significantly increase the computational resources required by topology losses, limiting their applicability.

\begin{figure*} 
  \centering
  \includegraphics[width=0.75\textwidth]{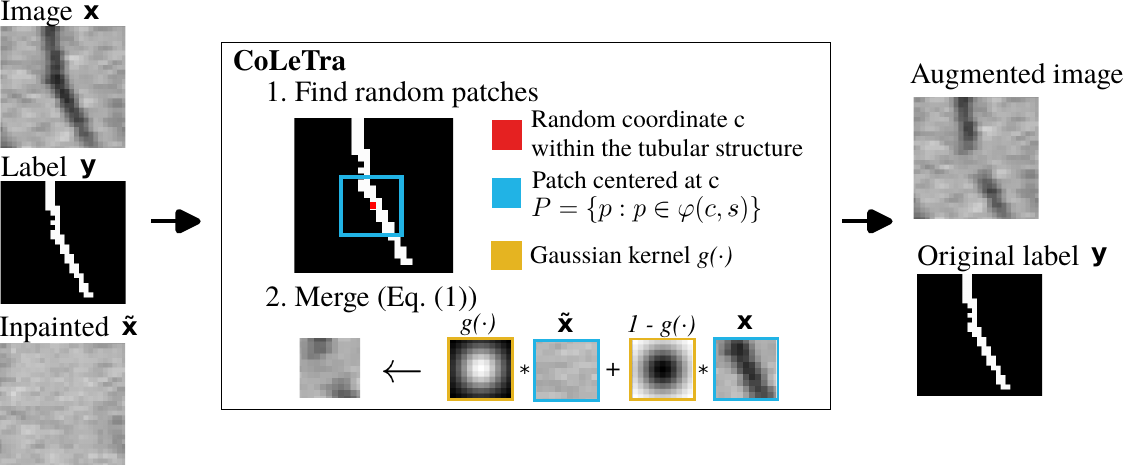}
  \caption{Overview of CoLeTra data augmentation.} \label{fig:coletra}
\end{figure*}

In this paper, we present CoLeTra, the first, to the best of our knowledge, data augmentation method for improving topology accuracy in image segmentation.
CoLeTra uses image inpainting to artificially create images with the appearance of disconnected tubular structures while keeping the ground truth labels as they are (see \Cref{fig:coletra}), teaching models that structures may be connected even if they appear disconnected.
We show with extensive experiments on four datasets, two architectures, and six loss functions that this simple strategy generally leads to segmentations with more accurate topology, \ie, segmentations with the number of connected components and holes more similar to the ground truth.
CoLeTra can be easily incorporated into existing training pipelines, and, importantly, it can be utilized jointly with other methods that consider topology, such as topology loss functions, as well as with other data augmentations.
Furthermore, CoLeTra adds only a negligible overhead during the optimization and, unlike other topology-enhancing strategies \cite{shit2021cldice,hu2019topology,hu2022structure}, it requires no extra GPU memory.
Specifically, our contributions are the following:

\begin{itemize}
    \item We present the first data augmentation method for image segmentation specifically designed to and demonstrated to improve topology accuracy.
    \item Our method improves topology accuracy across a wide range of settings, even when optimizing topology loss functions.
    \item We release a dataset ideally suited for evaluating image segmentation methods with an emphasis on topology accuracy. This dataset release aims to foster future research in the field.
\end{itemize}
\section{Related work}\label{sec:prev}
\textbf{Topology loss functions} 
Several topology loss functions aim at finding and improving the segmentation around \textit{critical points} (\ie, the pixels that change the topology of the segmentation to the desired one). At the core of most of these loss functions \cite{hu2019topology,gabrielsson2020topology,clough2020topological,shin2020deep,byrne2022persistent,he2023toposeg,oner2023persistent,stucki2023topologically}, we find persistence homology (PH) \cite{edelsbrunner2002topological}---a tool to analyze topological features.
Particularly in image segmentation, PH tracks how connected components, holes, and cavities appear and disappear as all possible thresholds are applied to softmax probability values \cite{hofer2017deep}.
Libraries, such as Gudhi \cite{maria2014gudhi}, facilitate the computation of PH and the finding of critical points that topology loss functions aim to correct.
There exist other topology loss functions that find critical points differently \cite{hu2022structure,liao2023segmentation}.
However, these and PH-based topology loss functions are very expensive to compute, forcing researchers to find critical pixels on small image patches, thus, disregarding global topology.
Our method, in contrast, increases computational time only by a small margin and, as we show, it improves global topology accuracy.

In parallel to this trend, other topology loss functions focus on improving the accuracy around the skeleton of tubular structures \cite{shit2021cldice,shi2024centerline,kirchhoff2024skeleton}.
These methods rely on skeletonization algorithms (and, some methods, on distance transforms) that must be computed constantly during training.
When computed on the CPU \cite{shi2024centerline,kirchhoff2024skeleton}, these methods increase running time, and, when computed on the GPU via min/max pooling \cite{shit2021cldice}, these methods increase GPU memory usage.
In contrast to these approaches, our method does not require extra GPU memory and it does not rely on computationally-expensive distance transform functions.

\noindent
\textbf{Prior knowledge via data augmentation} 
Data augmentation can integrate prior knowledge into the models during training by leveraging domain-specific knowledge.
In one example, a data augmentation method based on the appearance of local vessel intensity against the surrounding background was developed \cite{parella2024new}.
The medical knowledge that tumors can grow in different locations has been incorporated by randomly flipping the images during the optimization \cite{myronenko20193d}.
In another example, intra and inter-rater variability was simulated by applying random deformations to the segmentation mask boundaries \cite{javaid2019semantic}.
In other domains, \citet{yang2021mask2defect} proposed a data augmentation transformation to create defects in metal surfaces accounting for new defect types that only appeared in the test set.
Furthermore, recent work has shown that even standard data augmentation techniques can improve topology accuracy \cite{Valverde_2025_BMVC}.

\noindent
\textbf{Inpainting to increase generalizability} 
Image inpainting is used for reconstructing missing parts of an image, \eg, to replace a removed object. However, a range of other applications have emerged. Inpainting has been used as a self-supervised learning task to achieve models with better feature representations \cite{pathak2016context,zhong2020random,chen2019self,li2021learning}. 
Another line of research employs image inpainting to achieve larger, more diverse training sets \cite{ruiz2019anda,ruiz2020ida,zhang2020learning,wang2022creagan,nie2022synposes,saha2021data,he2024image}.
These approaches use off-the-shelf image inpainting methods to create ``background images'', where objects have been removed, for later to place other objects on those background images.
In this paper, we use image inpainting to augment data with appearance of change in topology.
\section{Method}
In this section, we present our \textbf{Co}ntinuity \textbf{Le}arning \textbf{Tra}nsformation (\textbf{CoLeTra}), a data augmentation method to improve topology accuracy.
Let $\mathbf{x} \in \mathbb{R}^{d}$ be an image of $d$ pixels from a training mini-batch, and let $\mathbf{y} \in \{0,1\}^{d}$ be its corresponding ground-truth segmentation mask.
CoLeTra transforms $\mathbf{x}$ during training:

\begin{equation} \label{eq:mergecoletra}
\begin{split}
    x_p \leftarrow & g(p,c,\sigma) \tilde{x}_p + (1 - g(p,c,\sigma)) x_p \\
    & \forall p \in \varphi(c,s), \forall c \in C,
\end{split}
\end{equation}
where $C = \{c_1, \dots ,c_n\}$ is a set of $n$ pixel coordinates such that $y_{c_i} = 1$; $\tilde{\mathbf{x}}$ is an inpainted version of $\mathbf{x}$ with all the thin structures removed;
$\varphi(c,s)$ is the set of pixel coordinates within a window size of $s \times s$ centered at $c$;
and $g(p,c,\sigma) = \exp(-\frac{1}{2} \frac{||p-c||^2_2}{\sigma^2}) \in [0,1]$ is the value of a Gaussian kernel that weighs the contribution of $\tilde{\mathbf{x}}$ and $\mathbf{x}$ at pixel $p$ (see \Cref{fig:coletra}).
For the sake of clarity, we simplified our notation; CoLeTra extends to multi-class problems, and, as we show, to multi-channel n-dimensional images.
Intuitively, CoLeTra erases random parts of the tubular structures via image inpainting (simulating low resolution/contrast) while leaving the ground truth as it is, teaching the models that structures may be connected even if they appear disconnected.

CoLeTra is agnostic to the inpainting method, allowing it to always utilize state-of-the-art inpainting methods and inpainting methods that are more suitable for specific datasets.
In preliminary experiments, we found that inpainting methods that lead to unrealistic, unnatural-looking disconnections \cite{telea2004image} were detrimental to CoLeTra.
In our implementation, we used LaMa \cite{suvorov2022resolution} to remove all tubular structures.
LaMa was used before the training to avoid on-the-fly inpainting, which would be costly and inefficient as LaMa always inpaints the images in the same way.
To ensure smooth borders and complete coverage of thin structures, the areas for inpainting were defined by dilating the labels three times:
\begin{equation}
    \mathbf{\tilde{x}} = LaMa(\mathbf{x}, dilate(\mathbf{y}, times=3)).
\end{equation}

CoLeTra uses a simple method for finding where to artificially break the thin structures.
During training, CoLeTra generates a set of random pixels by sampling, with the same probability, $n$ pixels containing the structure to be segmented:
\begin{equation}
C \subset \{ i : y_i = 1 \} : |C|=n.
\end{equation}
CoLeTra, then, centers the inpainted patches in each pixel belonging to the subset $C$.

CoLeTra's simplicity provides two important benefits.
First, CoLeTra only adds a negligible overhead (in the order of \textit{ms} per iteration) to the training while requiring no extra GPU memory.
Second, the strategy for finding where to disconnect structures can be tailored to each dataset based on prior knowledge.
For instance, disconnections could be applied to areas with lower contrast or at points where structures bifurcate, prioritizing the correction of regions with these specific characteristics.
In this work, we show in datasets with accurate and topologically inaccurate labels that, even with a simple strategy, CoLeTra achieves segmentations topologically more accurate.

\section{Experiments}

\subsection{Benchmark} \label{sec:mainresults}
\paragraph{Datasets and data split}
We evaluated CoLeTra on four datasets: DRIVE \cite{staal2004ridge}, Crack500 \cite{yang2019feature}, PulmonaryAV \cite{cheng2024fusion}, and the Narwhal dataset---that we make publicly available.
The DRIVE dataset comprised 20 fundus retina images collected for diabetic retinopathy detection, and the segmentation task focused on blood vessel segmentation.
Due to the small dataset size, we conducted three-fold cross validation, and we further applied a 0.8:0.2 split to yield the training and validation sets.
The Crack500 \cite{yang2019feature} dataset included 500 images with pavement cracks, for which we utilized the original training, validation, test set splits of 250, 50, and 200 images, respectively.
The PulmonaryAV  dataset included 106 3D CT scans of chest images that we divided into a 74-10-22 train-validation-test split.
In this dataset, the task was binary segmentation of pulmonary arteries and veins.
The Narwhal dataset comprised synchrotron 3D images from a narwhal tusk, where the task was to segment the dentine tubules (see details in \Cref{app:narwhal}).
The dataset consisted of 20 patches from a single scan for training, one patch from a different scan for validation, and three patches from other scans for testing.
Importantly, the training labels of DRIVE, Crack500, and PulmonaryAV are accurate, whereas the training labels of the Narwhal dataset are topologically incorrect since they were generated semi-automatically, resulting in thousands of broken structures.
The test set labels of the Narwhal dataset were obtained differently to maximize topology accuracy, allowing to measure topology accuracy reliably (\Cref{app:narwhal}).
Consequently, on the Narwhal dataset, while overlapping metrics such as the Dice coefficient can indicate overall segmentation quality, it allows to assess more effectively which methods produce topologically accurate segmentations.

\paragraph{Experimental setup}
We evaluated CoLeTra on 48 scenarios, combining two architectures, six loss functions, and the four datasets mentioned above (three datasets with topologically accurate training labels, and one with topologically inaccurate training labels).
We trained a DynUNet, which is based on the state-of-the-art nnUNet \cite{isensee2021nnu}, and an AttentionUNet \cite{oktay2018attention}, that incorporates attention-based modules.
The loss functions that these models optimized were selected based on their characteristics. 
Cross Entropy (CE) and Dice loss \cite{milletari2016v} are the most used loss functions for image segmentation; Dice loss, in addition, accounts for class imbalance.
RegionWise loss (RWLoss) \cite{valverde2023region} is a loss function that incorporates pixel importance with distance maps and can improve topology accuracy \cite{liu2024enhancing}.
clDice \cite{shit2021cldice}, Warping \cite{hu2022structure}, and Topoloss \cite{hu2019topology}, are topology loss functions; clDice focuses on achieving accurate skeletons in tubular structures; Warping loss focuses on rectifying critical pixels that affect the topology of the prediction; and Topoloss focuses on correcting topological features leveraging persistence homology.
For the DRIVE and Crack500 datasets, we trained 2D models, whereas for the PulmonaryAV and Narwhal datasets we trained 3D models.
For the 48 experimental scenarios, we compared models trained with extensive data augmentation (baseline) against the same setup plus our CoLeTra.
We repeated each experiment with three different random seeds.
Our experiments ran on a cluster with NVIDIAs V100 (32GB), with each experiment taking between 1 hour and 24 hours depending on the dataset and loss function.
Other details regarding the optimization can be found in \Cref{app:optimization}.

\paragraph{Data augmentation}
We employed extensive data augmentation in all experiments to ensure a realistic evaluation since, without it, the performance difference between methods could be artificially inflated in favor of CoLeTra.
All images were randomly augmented, with a 50\% probability in each transform independently, with random Gaussian noise ($\mu=0, \sigma=1$), random gamma correction ($\gamma \in [0.8, 1.2]$), and random axis flip.
DRIVE images were, in addition, randomly re-scaled between $[1, 1.75]$.
Since PulmonaryAV and Narwhal volumes were too big to fit the GPU memory, we sampled, after applying the Gaussian noise, 15 $128 \times 128 \times 128$ patches in both datasets.
We sampled only 15 patches to ensure diversity in the data augmentation; note that we optimized the models for a certain number of iterations and not epochs.

\paragraph{CoLeTra's hyper-parameters}
CoLeTra has two main hyper-parameters: the number of patches ($n$, \ie, $|C|$ in \Cref{eq:mergecoletra}) and patch size ($s$).
Their optimal value, as in any data augmentation method, depends on the dataset and task.
Since our goal was not to find their optimal value in each experimental scenario, we conducted a very simple strategy that explored only nine configurations per architecture and dataset.
We optimized Cross Entropy loss with patches of size 11, 15, and 19---since the structures have a similar width---and a number of patches such that, after applying CoLeTra, the maximum covered area of the structures corresponds to approximately 40\%, 50\%, and 60\%, avoiding breaking the structures too little or too much.
The exact configurations that we explored and the configuration that yielded the best performance are detailed in \Cref{app:coletrahyperparamopt}.

\paragraph{Metrics}
We measured the Dice coefficient, the centerline Dice (clDice) \cite{shit2021cldice}, the Betti errors, the false positive rate (FPR), and the Precision.
We chose these metrics since they measure different characteristics of the segmentations at the pixel and topology level.
The Dice coefficient is a standard metric used in image segmentation that measures the overlap between the ground truth and the prediction.
clDice measures performance around the center line of the structures \cite{shi2024centerline}.
The FPR and the Precision allows to elucidate whether and to what extent CoLeTra increases the number of false positives, which is important since CoLeTra is designed to connect structures that appear disconnected.
The Betti errors refer to the difference in the number of connected components ($\beta_{0e}$), holes ($\beta_{1e}$), and cavities ($\beta_{2e}$) between the predicted mask and the ground truth.
These can be computed locally through several random (or sliding window) patches across the image \cite{lin2023dtu}, or globally in the entire image depending on whether local or global topology is more relevant.
Here, we report the first global Betti error ($\beta_{0e}$) as it is the most relevant for all tasks, while the other Betti errors can be found in \Cref{app:bettierrors}.
Furthermore, we conducted non-parametric permutation tests with random 10000 iterations ($\alpha$=0.05) comparing the baseline methods against their CoLeTra-augmented counterparts.

\begin{figure*}
  \centering
  \includegraphics[width=0.85\textwidth]{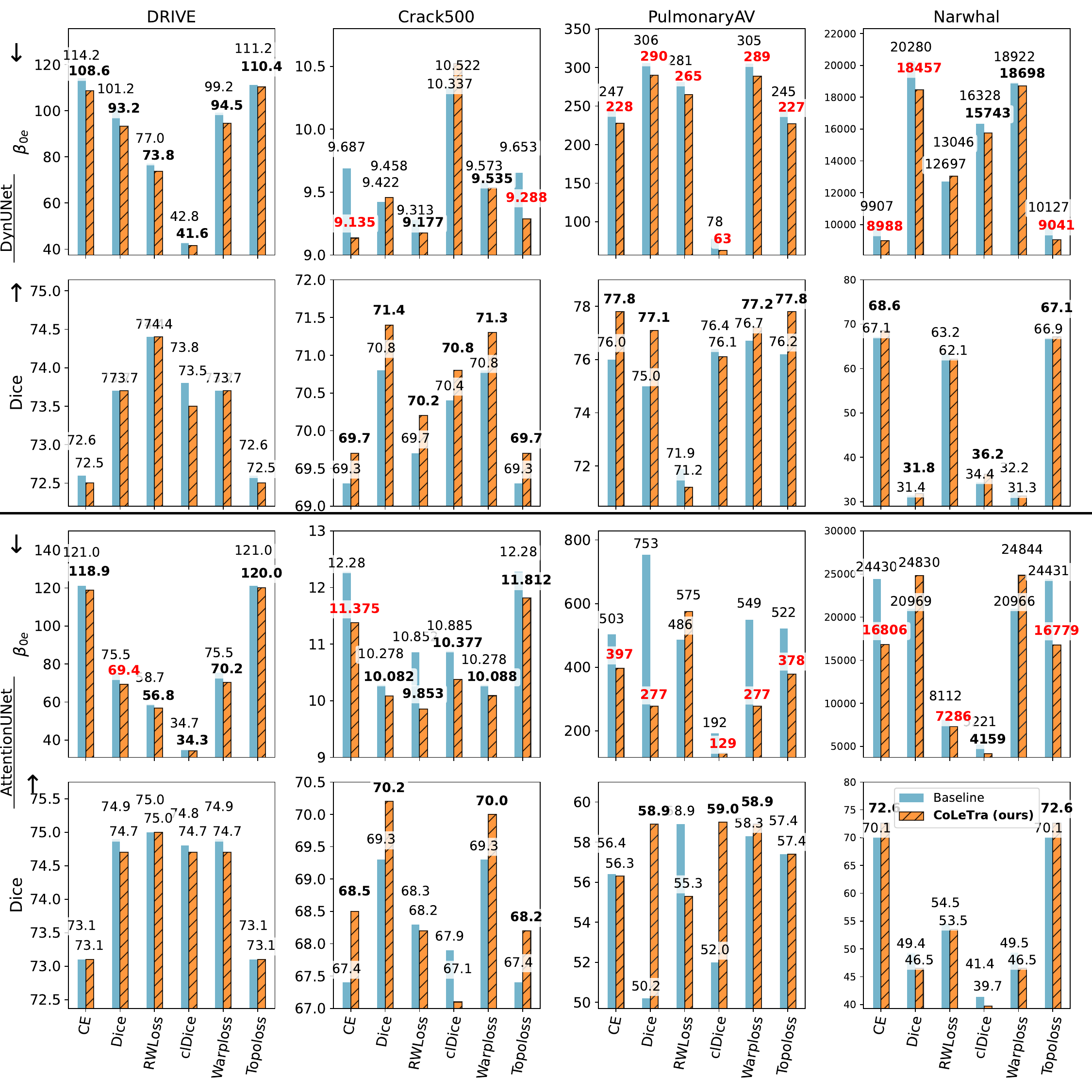}
  \caption{Comparison of baselines (four datasets, two architectures, six loss functions) with and without CoLeTra. \boldred{Red}: statistically significant improvement. Table with all performance metrics in \Cref{app:performance}.} \label{fig:results_barplot}
\end{figure*}

\begin{figure*}
  \centering
  \includegraphics[width=0.9\textwidth]{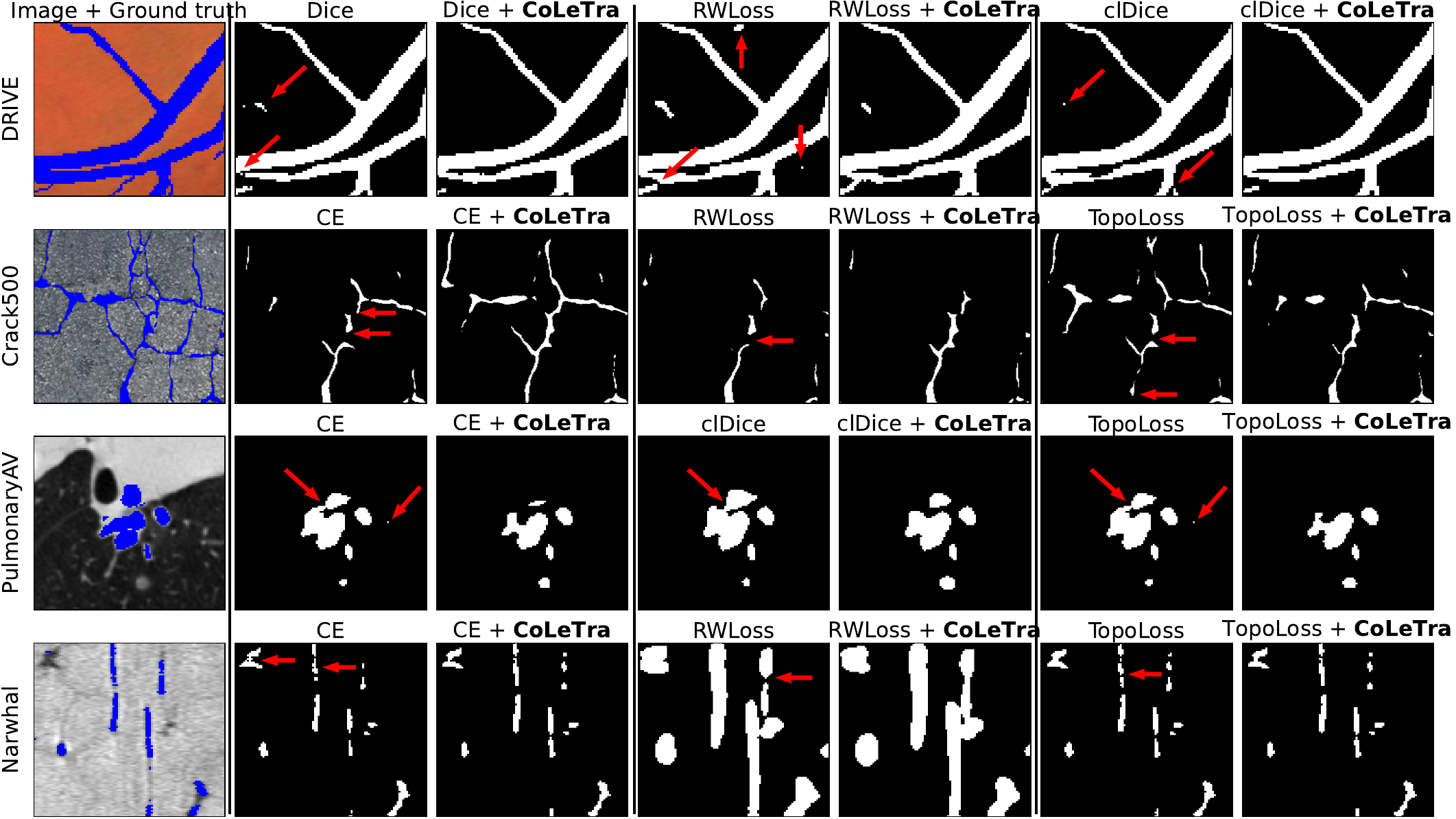}
  \caption{Example segmentations on four datasets and the best-performing loss functions in terms of Betti error with DynUNet architecture, with and without CoLeTra. Arrows: topological errors fixed by CoLeTra.} \label{fig:results_qualitative}
\end{figure*}

CoLeTra generally led to \textbf{smaller Betti errors} across all datasets, loss functions, and architectures, indicating higher topology accuracy (see \Cref{fig:results_barplot}, \Cref{fig:results_qualitative}, \Cref{app:performance}, \Cref{app:pulmonary}).
These improvements were observed even when optimizing topology loss functions: CoLeTra always improved Topoloss segmentations, and it improved clDice and Warping loss in 14/16 cases.
When optimizing non-topology loss functions (CE, Dice, and RWLoss), CoLeTra consistently yielded topologically more accurate segmentations on DRIVE dataset, and in the majority of the cases in Crack500, PulmonaryAV and the Narwhal dataset.
Notably, the smallest Betti errors across all four datasets were achieved with CoLeTra.
Moreover, CoLeTra did not appear to create false cycles or increase the number of false positives since the second global and local Betti errors ($\beta_{1e}$), the FPR, and the Precision generally did not deteriorate (\Cref{app:bettierrors,app:performance}).
Furthermore, in many cases, CoLeTra led to better clDice and Dice coefficients (\Cref{app:performance}).

\subsection{Sensitivity analysis}
We investigated how the two main hyper-parameters of CoLeTra---the size and number of patches---affect the topology accuracy gains.
To this end, we optimized DynUNet on the DRIVE dataset with the Dice loss as described in \Cref{sec:mainresults}.
In addition to measuring the Betti error, we counted the number of small ($<$ 30 pixels) connected components, as these structures represent fragments of broken blood vessels.
With the DRIVE images containing, on average, only one or two small connected components per image according to the ground truth, a small number of these connected components in the segmentation masks indicates high topology accuracy.
We explored the number of patches $n=\{10, 20, \dots, 200\}$ and the patch sizes $s=\{3, 7, \dots, 67\}$ in two separate sets of experiments.

CoLeTra decreased the Betti error and number of small connected components with patch sizes between 10 and 120 (see \Cref{fig:trends} (a,b)).
Beyond 120 patches, the Betti error and number of connected components oscillated.
Similarly, the patch sizes between $3^2$ and $19^2$ decreased the Betti error and the number of small connected components (see \Cref{fig:trends} (c,d).
Thereafter, these metrics also started to oscillate.
\Cref{fig:trends} also shows that a moderately different patch size and number of patches from the ones we used in \Cref{sec:mainresults} ($n=63, s=15$) leads to similar topology accuracy.

\section{Discussion}

We introduced CoLeTra, a data augmentation strategy for improving topology accuracy by incorporating the prior knowledge that structures that appear disconnected can be connected.
We evaluated CoLeTra on four benchmark datasets with different architectures and loss functions, and we studied its sensitivity to several hyper-parameter choices.

CoLeTra \textbf{consistently} improved topology accuracy across different architectures and loss functions on datasets with topologically accurate (DRIVE, Crack500, PulmonaryAV) and  inaccurate training labels (Narwhal).
CoLeTra not only reduced the Betti errors but, in many cases, it also improved clDice and Dice coefficient while did not generally increased the number of false connections or false positives (see \Cref{app:performance}).
It is important to note, however, that while improvements in non-topology metrics are a beneficial side effect, they are not the primary focus of CoLeTra.
Instead, CoLeTra emphasizes topology accuracy, which is critical for downstream quantification tasks, such as counting the number of connected components.
In some experimental scenarios, the performance gains were moderate, which aligns with the expected outcome of adding just one extra data augmentation transformation on top of several other data augmentations during training.
A single data augmentation technique alone typically does not lead to very large performance improvements \cite{nanni2021comparison,mumuni2022data}---even advanced strategies, such as CutOut and MixUp (see \citet{yun2019cutmix}).

CoLeTra demonstrates, for the first time, that topology can be improved via data augmentation.
Although previous works have also used image inpainting to increase datasets' diversity (see \Cref{sec:prev}), CoLeTra is unique in that it is specifically designed to improve topology accuracy and it demonstrates to do so, even when optimizing topology loss functions.
CoLeTra's improvements demonstrate that data augmentation methods can be used orthogonally and complement topology loss functions and, potentially, deep learning architectures and postprocessing methods that focus on improving topology accuracy (\eg, \cite{jin2019dunet,yang2022dcu,qi2023dynamic,sophie2024restoring}).
This capability to enhance topology accuracy alongside other methods during training is important since typical deep learning methods cannot guarantee accurate topology unless the topological features are known beforehand \cite{clough2020topological}, which does not often occur.
Moreover, CoLeTra was particularly useful when the dataset training labels were topologically incorrect; in the Narwhal dataset, CE and Topoloss achieved the best segmentations according to the standard metrics (Dice, clDice) and, after applying CoLeTra, topology accuracy (the Betti error) further improved.

CoLeTra's simplicity ensures that CoLeTra is fast, while requiring no extra GPU memory.
This is crucial because it offers the possibility to improve topology accuracy in scenarios where other topology-enhancing methods are limited by their computational cost \cite{hu2019topology,hu2022structure,liao2023segmentation} or larger GPU memory requirements \cite{shit2021cldice}, and because an otherwise complex, computationally-expensive data augmentation would not be practically useful for iterative training.
Additionally, CoLeTra's simple strategy to determine where to break the tubular structures can be extended to focus on fixing specific type of breaks \cite{ren2024self} or in other locations considered important.
Exploring performance gains in particular cases was beyond the scope of our study; instead, we demonstrated that even a simple approach can improve topology accuracy.

\begin{figure} 
  \centering
  \includegraphics[width=0.45\textwidth]{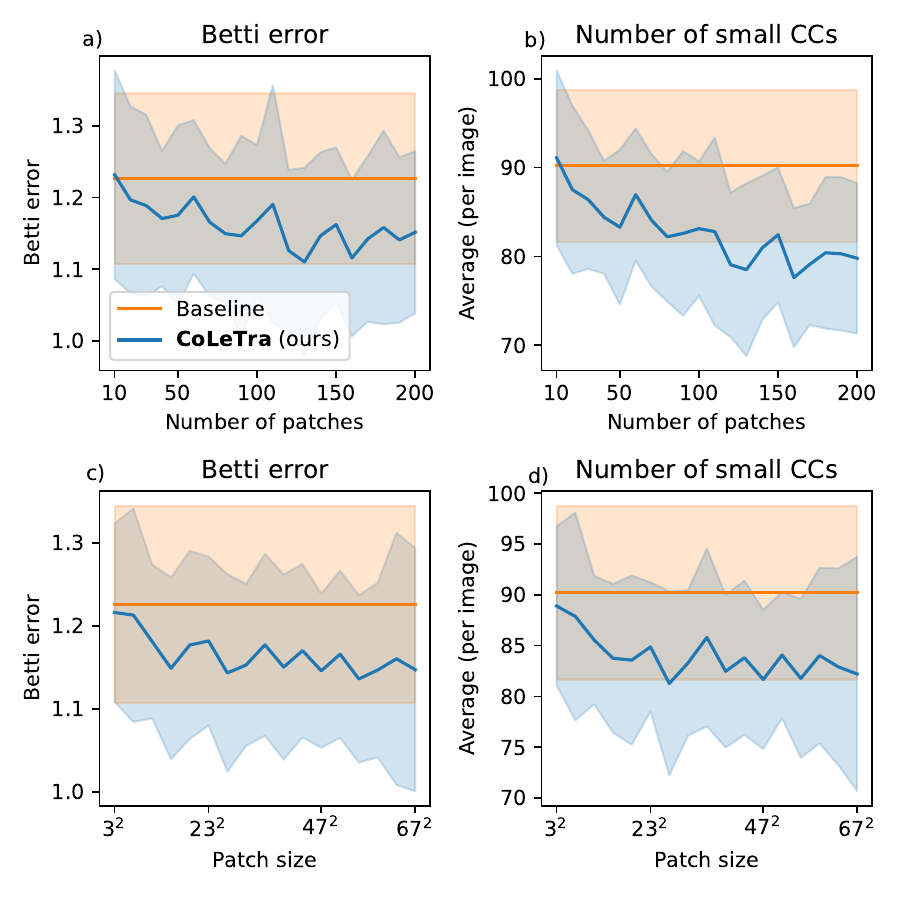}
  \caption{Betti error and number of small ($<$ 30 pixels) connected components achieved with CoLeTra on the DRIVE dataset with different number of patches (a,b) and patch sizes (c,d). (CC: connected components)} \label{fig:trends}
\end{figure}

CoLeTra has two main hyper-parameters: the size, and the number of patches that will disconnect the tubular structures.
In our comparisons across architectures, datasets, and loss functions, we utilized the hyper-parameter values that we found on a very small grid search---likely a sub-optimal configuration that, nevertheless, increased the topology accuracy of the segmentations.
More accurate segmentations could be achieved with a more exhaustive grid search, and by jointly accounting for the datasets' particularities, such as the disproportion between the pixels from thin and thick structures \cite{liew2021deep}, class imbalance, and noisy labels \cite{song2022learning}.

Our sensitivity analysis showed that a large number of patches and a large patch size tend to decrease the Betti error and the number of small connected components, indicating higher topology accuracy.
The range of patch sizes that improved topology accuracy (from $3^2$ to $23^2$, \Cref{fig:trends} (c,d)) overlaps with the range of the widths of the blood vessels (from 2 to 18.5), demonstrating that patch size can be tuned, intuitively, based on the structures' width.
Regarding the number of patches, we hypothesize that the optimal value depends on the accuracy of the annotations.
In datasets with topologically inaccurate labels marking visually-connected structures as disconnected (see \Cref{fig:teaser} (Narwhal)), the optimization will reinforce such incorrect disconnection.
Since CoLeTra promotes the opposite (\ie, connecting visually-disconnected structures), a large number of patches may help compensating the structures mislabeled as disconnected.

\section{Conclusion}
CoLeTra increases topology accuracy in a wide range of settings, including different architectures, loss functions, and datasets with topologically accurate and inaccurate labels.
Additionally, it does not deteriorate---often, it improves---other metrics, such as Dice coefficient and clDice.
Our sensitivity analysis showed that CoLeTra is robust to different hyper-parameter choices, and these hyper-parameters are largely intuitive, facilitating their tuning.
Finally, to further encourage research in the domain of image segmentation and topology, we release the Narwhal dataset.

\paragraph{Limitations} The advantageousness of CoLeTra, as any other data augmentation, will depend on the dataset and task.
CoLeTra's advantageous capability of promoting the connectivity between visually-disconnected structures may be disadvantageous on datasets with visually-disconnected structures that should remain disconnected and when boundary noise in the labels is high (e.g., Crack500 dataset).
CoLeTra assumes that false positives (\ie, incorrectly connected structures) are relatively rare in the training labels, i.e., that the training labels indicate entire and/or broken structures.
Additionally, our hyperparameter sensitivity analysis was restricted to DynUNet on DRIVE.

\section*{Acknowledgments}
This work was supported by Villum Foundation and NordForsk.

\section*{LLM disclosure}
We used LLMs/AI tools to improve readability in certain sentences of the manuscript.

LLMs/AI tools were \textbf{not} used: 1) to generate any text from scratch, 2) to generate any code, 3) to generate any figure or table, 4) to come up with the research question, 5) to conduct literature research, 6) to design the method, 7) to identify, find, download and preprocess any dataset, 8) to identify, find, download and prepare the methods compared to ours.
{
    \small
    \bibliographystyle{ieeenat_fullname}
    \bibliography{main}
}

\clearpage
\appendix
\maketitlesupplementary

\section{The Narwhal dataset} \label{app:narwhal}
\subsection{Image acquisition and reconstruction}
Narwhal tusks, composed of an inner dentine ring with pores (dentine tubules) running through it \cite{zanette2015ptychographic}, can be effectively studied using microcomputed tomography, particularly synchrotron radiation microcomputed tomography (SR-µCT) \cite{wittig2022opportunities}. In order to study its internal structure, a piece of narwhal tusk was cut radially, with respect to the tusk length, to a square rod with side lengths 460 µm by 530 µm using a diamond blade saw (Accutom-5, Struers, Ballerup, Denmark).
The rod was imaged with SR-µCT in the Paul Scherrer Institue, Viligen, Switzerland, at the TOMCAT beamline X02DA of the Swiss Light Source.
Several scans were measured along the long axis of the rod to image it entirely.
For each scan, 2000 projections were collected covering 180° with an energy of 18 keV, and an exposure time of 150 ms.
30 closed beam images were used for dark field correction and 50 open beam images for flat field correction.
A 20 µm LuAG:Ce scintillator, a microscope (Optique Peter, France) with a $\times$20 magnification, and a  PCO.Edge 5.5 camera (PCO AG, Kelheim, Germany) gave rise to an isotropic voxel size of 0.325 µm.

Image reconstruction was performed on-site using the GridRec algorithm in the RecoManagerRa pipeline plug-in for ImageJ.
In the reconstruction, a Parzen filter with a cut-off frequency of 0.5 was employed as well as standard ring removal.
The data was scaled to be between $-4*10^{-4}$ and $8*10^{-4}$, and output as 16-bit grey level images.
Each scan had dimensions of 2560 $\times$ 2560 $\times$ 2160 voxels.

\subsection{Training, validation, and test splits}
For the purpose of this study, we employed five scans containing abundant dentine tubules.
One scan, used for training, was divided into 64 overlapping patches.
Then, we discarded those patches containing background area, leading to our training set of 20 patches of, approximately, 591 $\times$ 530 $\times$ 583 voxels.
The validation set corresponds to a 500 $\times$ 500 $\times$ 500 patch from another scan, and the test set is comprised by three 500 $\times$ 500 $\times$ 500 patches, each from a different scan.

\subsection{Labels}
Due to the extremely large size of the images, making their manual annotation prohibitively expensive, we derived pseudo-labels for the training set and pseudo-ground truths for the validation and test set.
The pseudo-labels were obtained via thresholding to simulate the common practice employed by researchers when analyzing these type of images.
The pseudo-ground truths, used exclusively to evaluate the predictions, were derived through a more elaborated process yielding segmentation masks that, although they were not perfect, they permit measuring topology accuracy reliably (i.e., counting the number of tubules), and permit quantifying tubules' properties, such as their directionality.

\paragraph{Pseudo-labels (training set)} First, we thresholded the images and considered as ``dentine tubule'' those voxels with intensities below 17,450---threshold that we found experimentally. Then, we removed non-tubular areas that we identified with a distance transform.

\paragraph{Pseudo-ground truth (validation and test set)} The pseudo-ground truth labels are created by first tracking the tubules through the volume and then assigning labels to the tracks. In unmineralized regions of the tusk, the intensity of the tubule and the regions are the same, which makes segmentation challenging. By utilizing that the dentine tubules are running predominantly along the $z$-direction of the volume, we can track the tubules to obtain their center lines. 

\begin{figure*} 
  \centering
  \includegraphics[width=1\textwidth]{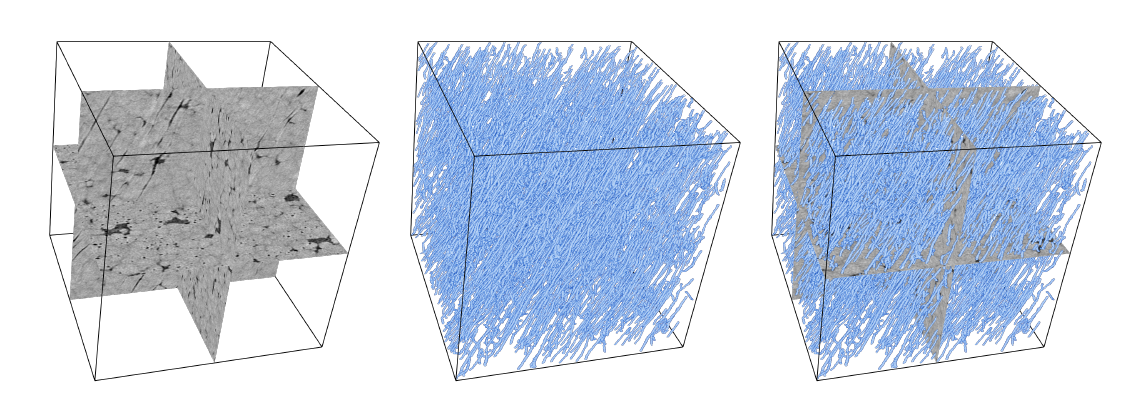}
  \caption{Example of labels in the Narwhal dataset. Left shows orthogonal central slices from the volume, middle shows the labels, and right shows the two together.} \label{fig:narwhale_example}
\end{figure*}

Tracking is done by initially detecting the tubule's center position and connecting detected points moving along the $z$-direction. The center position is detected as the local maxima in a Gaussian-smoothed image ($\sigma = 1.5$). We connect mutually closest points in consecutive frames under some constraints. Points in one frame must be separated by an Euclidean distance of at least five voxels, be within an Euclidean distance of no more than three voxels in consecutive frames, if no points are detected within five frames tracks are stopped, and we only keep tracks that are at least 20 voxels long (Euclidean distance). Based on the tracks, we create a label image containing a thresholded version of the original image (threshold value of 30,720) but masked along the tracks. We employ a minimum mask of a $3 \times 3$ window that the label should minimally be, and a maximum mask of a circle with a diameter of $5 \times 5$ that the label can maximally obtain. If the thresholded image is within these limits, the threshold will be chosen. After labeling, we see tubules being tracked nicely through the unmineralized regions, and labels that connect in a complex network as you will expect for dentine tubules. Tracks are illustrated in \Cref{fig:narwhale_example}.

\subsection{Dataset availability}
The Narwhal dataset, released under CC BY-NC 4.0 licence, can be downloaded at \url{https://archive.compute.dtu.dk/files/public/projects/CoLeTra_Narwhal}.

\section{Optimization} \label{app:optimization}
\subsection{Architectures}
We optimized two model architectures: DynUNet\footnote{https://docs.monai.io/en/stable/networks.html\#dynunet} and AttentionUNet\footnote{https://docs.monai.io/en/stable/networks.html\#attentionunet}.

\textbf{DynUNet} had the same architecture in all our experiments, differing only in the dimension of the kernels and the input channels.
In DRIVE and Crack500 datasets we used 2D kernels ($spatial\_dims=2$), and in PulmonaryAV and the Narwhal dataset we used 3D kernels ($spatial\_dims=3$).
DRIVE and Crack500 contained RGB images ($in\_channels=3$) whereas PulmonaryAV and the Narwhal dataset contained single-channel images ($in\_channels=1$).
DynUNet had four downsampling blocks that used convolutions with strides $=2$ to reduce the dimensionality of the features.
Additionally, DynUNet employed deep supervision, optimizing the output of three upsampling blocks.
The exact configuration of DynUNet was:

\begin{lstlisting}
DynUNet(spatial_dims, in_channels,
    out_channels=2,
    kernel_size=[3, 3, 3, 3, 3, 3],
    strides=[1, 2, 2, 2, 2, 2],
    upsample_kernel_size=[2, 2, 2, 2, 2],
    deep_supervision=True,
    deep_supr_num=3)
\end{lstlisting}

\textbf{AttentionUNet} architecture differed across datasets.
We initially aimed to configure AttentionUNet to match, as much as possible, the number of parameters of DynUNet. However, due to the larger GPU memory requirements, we had to use a smaller and shallower architecture in some datasets.
The dimension of the kernels and the input channels were as in DynUNet.
Additionally, the depth and the number of feature maps per block were as follows:

\underline{DRIVE} and \underline{Crack500}
\begin{lstlisting}
AttentionUNet(spatial_dims=2,
    in_channels=3, out_channels=2,
    channels=[32, 64, 128, 256, 512, 512],
    strides=[1, 2, 2, 2, 2, 2])
\end{lstlisting}

\underline{PulmonaryAV} and \underline{Narwhal}
\begin{lstlisting}
AttentionUNet(spatial_dims=3,
    in_channels=1, out_channels=2,
    channels=[16, 24, 32],
    strides=[1, 2, 2])
\end{lstlisting}

\subsection{Optimization}
We utilized Adam optimizer with a starting learning rate of 0.001 and weight decay of $10^{-5}$.
We decreased the learning rate during training: $\eta \leftarrow \eta (1 - \frac{iteration}{max\_iterations})^{0.9}$.
For DRIVE, Crack500, PulmonaryAV, and the Narwhal dataset, we optimized the models for 5000, 7000, 7000, 1000 iterations, respectively.
In addition, due to the class imbalance in the Narwhal dataset, we used weighted Cross entropy (including in Topoloss that relies on Cross entropy) with the following weights: $[0.05, 0.95]$.
The rectified region wise maps employed by RWLoss in the Narwhal dataset were also multiplied by those weights accordingly.

\section{CoLeTra's hyper-parameters search} \label{app:coletrahyperparamopt}
For each dataset and architecture, we explored a maximum of nine different configurations combining three patch sizes ($s$) and three number of patches ($n$).
For the patch sizes, we explored the values 11, 15, and 19 since the structures to be segmented have a width around that range.
For the number of patches, to avoid disconnecting too many or too few structures, we designed a simple strategy based on the number of pixels in the annotated areas.
Given the average number of pixels and a patch size, consider the number of patches such that the maximum covered area of the structures after applying CoLeTra corresponds to approximately 40\%, 50\%, and 60\% of the total.

\Cref{tab:opt_drive,tab:opt_crack500,tab:opt_pulmonary,tab:opt_narwhal} specify the combinations of hyper-parameters that were explored, and which one yielded the best performance in DynUNet (D) and AttentionUNet (A) architectures. Note that, in the Narwhal dataset, a single patch size of $15 \times 15 \times 15$ and $19 \times 19 \times 19$ already covers more than the expected number voxels from the structure to be found in the $128 \times 128 \times 128$ patches used during training. Thus, we only explored to apply a single patch when its size was 15 or 19.

\begin{table}
\begin{minipage}[b]{0.5\linewidth}
\centering
\caption{DRIVE}  \label{tab:opt_drive}
  \begin{tabular}{l|r|l} 
     $s$ & $n$ & \checkmark? \\
    \hline
    11 &  94 &  \\
    \hline
    11 &  117 &    \\
    \hline
    11 &  141 &  \\
    \hline
    15 & 50  &  \\
    \hline
    15 & 63  &  D  \\
    \hline
    15 &  75 &  \\
    \hline
    19 &  31 &  \\
    \hline
    19 &  39 &  A  \\
    \hline
    19 &  47 &  \\
    \hline
  \end{tabular} 
  
\end{minipage}\hfill
\begin{minipage}[b]{0.5\linewidth}
\centering
\caption{Crack500} \label{tab:opt_crack500}
  \begin{tabular}{l|r|l} 
     $s$ & $n$ & \checkmark? \\
    \hline
    11 &  57 &  \\
    \hline
    11 &  71 &    \\
    \hline
    11 &  86 &  \\
    \hline
    15 &  30 & A \\
    \hline
    15 & 38  &    \\
    \hline
    15 &  46 &  \\
    \hline
    19 &  19 & D \\
    \hline
    19 & 24  &    \\
    \hline
    19 &  28 &  \\
    \hline
  \end{tabular}
\end{minipage}
\end{table}

\begin{table}
\begin{minipage}[b]{0.5\linewidth}
\centering
    \caption{PulmonaryAV}  \label{tab:opt_pulmonary}
  \begin{tabular}{l|r|l} 
     $s$ & $n$ & \checkmark? \\
    \hline
    11 &  10 &  \\
    \hline
    11 &  13 &    \\
    \hline
    11 &  15 &  \\
    \hline
    15 &  4 &  \\
    \hline
    15 & 5  &    \\
    \hline
    15 &  6 & A \\
    \hline
    19 &  2 & D \\
    \hline
    19 & 3  &    \\
    \hline
  \end{tabular}
\end{minipage}\hfill
\begin{minipage}[b]{0.5\linewidth}
\centering
    \caption{Narwhal}  \label{tab:opt_narwhal}
  \begin{tabular}{l|r|l} 
     $s$ & $n$ & \checkmark? \\
    \hline
    11 & 2  & D \\
    \hline
    11 &  3 &    \\
    \hline
    11 & 4  &  \\
    \hline
    15 &  1 & A \\
    \hline
    19 & 1  &  \\
    \hline
  \end{tabular}
\end{minipage}
\end{table}

\section{Betti Errors} \label{app:bettierrors}
\Cref{tab:bettidyn,tab:bettiatt} show all the Betti errors (global and local) obtained in our experiments, which were computed via the Gudhi library \cite{maria2014gudhi}.
The global Betti errors were the difference in the Betti numbers between the entire ground-truth annotation and the prediction.
The local Betti errors were computed similarly but in patches of size 64 $\times$ 64 patches (2D images) or 48 $\times$ 48 $\times$ 48 (3D images).
The patches were extracted either randomly or in a sliding-window manner depending on whether the total number of non-overlapping patches that could be extracted from the images were more than 500 (for 2D images) or 100 (for 3D images).

\begin{table*}
\caption{Global and local Betti Errors (DynUNet)}
\label{tab:bettidyn}
\centering
\tiny{
\begin{tabular}{lllllllll}
\toprule
\multicolumn{3}{c}{} & $\beta_{0e}$ (global) & $\beta_{1e}$ (global)  & $\beta_{2e}$ (global) & $\beta_{0e}$ (local) & $\beta_{1e}$ (local)  & $\beta_{2e}$ (local)\\
\cmidrule(r){4-9}
\parbox[t]{2mm}{\multirow{12}{*}{\rotatebox[origin=c]{90}{\shortstack[c]{DRIVE}}}} & CE & baseline & 114.2 (11.8) & 35.5 (2.4) & --- & 1.3 (0.1) & 0.2 (0.0) & ---\\
 &  & +\textbf{CoLeTra} & \textbf{108.6} (10.2) & \textbf{33.6} (2.9) & --- & \textbf{1.3} (0.1) & 0.2 (0.0) & ---\\
\cmidrule(r){2-9}
 & Dice & baseline & 101.2 (10.8) & 28.5 (3.3) & --- & 1.2 (0.1) & 0.2 (0.0) & ---\\
 &  & +\textbf{CoLeTra} & \textbf{93.2} (9.2) & \textbf{26.2} (3.6) & --- & \textbf{1.1} (0.1) & 0.2 (0.0) & ---\\
\cmidrule(r){2-9}
 & RWLoss & baseline & 77.0 (8.2) & 29.1 (2.8) & --- & 1.0 (0.1) & 0.2 (0.0) & ---\\
 &  & +\textbf{CoLeTra} & \textbf{73.8} (6.2) & \textbf{28.2} (3.1) & --- & \textbf{1.0} (0.1) & 0.2 (0.0) & ---\\
\cmidrule(r){2-9}
 & clDice & baseline & 42.8 (7.3) & 24.0 (3.2) & --- & 0.7 (0.1) & 0.2 (0.0) & ---\\
 &  & +\textbf{CoLeTra} & \textbf{41.6} (6.6) & \textbf{22.1} (3.4) & --- & \textbf{0.7} (0.1) & 0.2 (0.0) & ---\\
\cmidrule(r){2-9}
 & Warploss & baseline & 99.2 (9.2) & 28.0 (3.8) & --- & 1.2 (0.1) & 0.2 (0.0) & ---\\
 &  & +\textbf{CoLeTra} & \textbf{94.5} (10.0) & \textbf{25.7} (2.8) & --- & \textbf{1.2} (0.1) & 0.2 (0.0) & ---\\
\cmidrule(r){2-9}
 & Topoloss & baseline & 111.2 (13.4) & 35.0 (2.3) & --- & 1.3 (0.1) & 0.2 (0.0) & ---\\
 &  & +\textbf{CoLeTra} & \textbf{110.4} (11.3) & \textbf{33.6} (2.8) & --- & \textbf{1.3} (0.1) & 0.2 (0.0) & ---\\
\midrule
\parbox[t]{2mm}{\multirow{12}{*}{\rotatebox[origin=c]{90}{\shortstack[c]{Crack500}}}} & CE & baseline & 9.687 (2.613) & 3.21 (1.489) & --- & 0.136 (0.019) & 0.031 (0.011) & ---\\
 &  & +\textbf{CoLeTra} & \textbf{9.135} (2.385) & \textbf{2.943} (1.103) & --- & \textbf{0.132} (0.017) & \textbf{0.029} (0.008) & ---\\
\cmidrule(r){2-9}
 & Dice & baseline & 9.422 (1.86) & 2.648 (0.453) & --- & 0.125 (0.014) & 0.023 (0.003) & ---\\
 &  & +\textbf{CoLeTra} & 9.458 (1.662) & \textbf{2.642} (0.488) & --- & \textbf{0.124} (0.012) & \textbf{0.022} (0.003) & ---\\
\cmidrule(r){2-9}
 & RWLoss & baseline & 9.313 (1.785) & 2.77 (0.376) & --- & 0.125 (0.012) & 0.021 (0.002) & ---\\
 &  & +\textbf{CoLeTra} & \textbf{9.177} (1.581) & 2.773 (0.345) & --- & \textbf{0.122} (0.013) & 0.021 (0.002) & ---\\
\cmidrule(r){2-9}
 & clDice & baseline & 10.337 (0.959) & 54.492 (49.838) & --- & 0.126 (0.014) & 0.378 (0.338) & ---\\
 &  & +\textbf{CoLeTra} & 10.522 (0.886) & 96.082 (44.202) & --- & 0.128 (0.011) & 0.667 (0.306) & ---\\
\cmidrule(r){2-9}
 & Warploss & baseline & 9.573 (1.757) & 2.655 (0.525) & --- & 0.125 (0.013) & 0.023 (0.003) & ---\\
 &  & +\textbf{CoLeTra} & \textbf{9.535} (1.694) & \textbf{2.608} (0.454) & --- & \textbf{0.124} (0.012) & \textbf{0.022} (0.003) & ---\\
\cmidrule(r){2-9}
 & Topoloss & baseline & 9.653 (2.351) & 3.2 (1.418) & --- & 0.135 (0.018) & 0.031 (0.011) & ---\\
 &  & +\textbf{CoLeTra} & \textbf{9.288} (2.357) & \textbf{2.957} (1.277) & --- & \textbf{0.133} (0.018) & \textbf{0.029} (0.01) & ---\\
\midrule
\parbox[t]{2mm}{\multirow{12}{*}{\rotatebox[origin=c]{90}{\shortstack[c]{PulmonaryVA}}}} & CE & baseline & 247 (14) & 26 (5) & 14 (0) & 17 (0) & 4 (0) & 3 (0)\\
 &  & +\textbf{CoLeTra} & \textbf{228} (15) & 27 (4) & 15 (0) & \textbf{15} (1) & 4 (0) & 3 (0)\\
\cmidrule(r){2-9}
 & Dice & baseline & 306 (27) & 24 (5) & 13 (1) & 18 (1) & 4 (0) & 3 (0)\\
 &  & +\textbf{CoLeTra} & \textbf{290} (25) & 25 (2) & 14 (0) & \textbf{17} (0) & 4 (0) & 3 (0)\\
\cmidrule(r){2-9}
 & RWLoss & baseline & 281 (26) & 25 (4) & 14 (1) & 17 (1) & 4 (0) & 3 (0)\\
 &  & +\textbf{CoLeTra} & \textbf{265} (21) & 26 (5) & 15 (0) & \textbf{17} (0) & \textbf{4} (0) & 3 (0)\\
\cmidrule(r){2-9}
 & clDice & baseline & 78 (12) & 24 (4) & 14 (1) & 5 (0) & 4 (0) & 3 (0)\\
 &  & +\textbf{CoLeTra} & \textbf{63} (6) & 25 (2) & 14 (0) & \textbf{5} (0) & 4 (0) & \textbf{3} (0)\\
\cmidrule(r){2-9}
 & Warploss & baseline & 305 (31) & 24 (4) & 14 (1) & 18 (1) & 4 (0) & 3 (0)\\
 &  & +\textbf{CoLeTra} & \textbf{289} (26) & 24 (3) & 14 (0) & \textbf{17} (1) & 4 (0) & 3 (0)\\
\cmidrule(r){2-9}
 & Topoloss & baseline & 245 (13) & 25 (4) & 14 (1) & 16 (0) & 4 (0) & 3 (0)\\
 &  & +\textbf{CoLeTra} & \textbf{227} (15) & 27 (4) & 14 (0) & \textbf{15} (1) & 4 (0) & 3 (0)\\
\midrule
\parbox[t]{2mm}{\multirow{12}{*}{\rotatebox[origin=c]{90}{\shortstack[c]{Narwhal}}}} & CE & baseline & 9907 (1128) & 336 (367) & 45 (16) & 56 (6) & 5 (0) & 0 (0)\\
 &  & +\textbf{CoLeTra} & \textbf{8988} (429) & 665 (288) & \textbf{36} (14) & \textbf{50} (3) & 5 (1) & \textbf{0} (0)\\
\cmidrule(r){2-9}
 & Dice & baseline & 20280 (249) & 1105 (25) & 51 (0) & 112 (3) & 7 (0) & 0 (0)\\
 &  & +\textbf{CoLeTra} & \textbf{18457} (1220) & \textbf{1103} (33) & 51 (0) & \textbf{100} (7) & \textbf{7} (0) & 0 (0)\\
\cmidrule(r){2-9}
 & RWLoss & baseline & 12697 (2712) & 806 (30) & 51 (0) & 73 (17) & 5 (0) & 0 (0)\\
 &  & +\textbf{CoLeTra} & 13046 (3102) & \textbf{707} (206) & \textbf{51} (0) & 76 (19) & \textbf{5} (0) & 0 (0)\\
\cmidrule(r){2-9}
 & clDice & baseline & 16328 (1693) & 1146 (6) & 51 (0) & 87 (9) & 7 (0) & 0 (0)\\
 &  & +\textbf{CoLeTra} & \textbf{15743} (592) & \textbf{1139} (12) & 51 (0) & \textbf{84} (2) & \textbf{7} (0) & 0 (0)\\
\cmidrule(r){2-9}
 & Warploss & baseline & 18922 (1388) & 1103 (25) & 51 (0) & 103 (9) & 7 (0) & 0 (0)\\
 &  & +\textbf{CoLeTra} & \textbf{18698} (781) & \textbf{1103} (35) & 51 (0) & \textbf{102} (4) & \textbf{7} (0) & 0 (0)\\
\cmidrule(r){2-9}
 & Topoloss & baseline & 10127 (1244) & 416 (240) & 70 (45) & 57 (8) & 5 (1) & 0 (0)\\
 &  & +\textbf{CoLeTra} & \textbf{9041} (682) & \textbf{414} (232) & \textbf{33} (23) & \textbf{50} (6) & \textbf{5} (1) & \textbf{0} (0)\\
\bottomrule
\end{tabular}
}
\end{table*}

\begin{table*}
\caption{Global and local Betti Errors (AttentionUNet)}
\label{tab:bettiatt}
\centering
\tiny{
\begin{tabular}{lllllllll}
\toprule
\multicolumn{3}{c}{} & $\beta_{0e}$ (global) & $\beta_{1e}$ (global)  & $\beta_{2e}$ (global) & $\beta_{0e}$ (local) & $\beta_{1e}$ (local)  & $\beta_{2e}$ (local)\\
\cmidrule(r){4-9}
\parbox[t]{2mm}{\multirow{12}{*}{\rotatebox[origin=c]{90}{\shortstack[c]{DRIVE}}}} & CE & baseline & 121.0 (13.6) & 34.9 (3.1) & --- & 1.4 (0.1) & 0.2 (0.0) & ---\\
 &  & +\textbf{CoLeTra} & \textbf{118.9} (13.7) & \textbf{32.2} (2.8) & --- & \textbf{1.4} (0.2) & 0.2 (0.0) & ---\\
\cmidrule(r){2-9}
 & Dice & baseline & 75.5 (19.4) & 29.2 (3.4) & --- & 1.0 (0.2) & 0.2 (0.0) & ---\\
 &  & +\textbf{CoLeTra} & \textbf{69.4} (16.7) & \textbf{27.8} (3.4) & --- & \textbf{0.9} (0.2) & 0.2 (0.0) & ---\\
\cmidrule(r){2-9}
 & RWLoss & baseline & 58.7 (11.7) & 29.9 (3.1) & --- & 0.8 (0.1) & 0.2 (0.0) & ---\\
 &  & +\textbf{CoLeTra} & \textbf{56.8} (11.2) & 29.9 (3.0) & --- & \textbf{0.8} (0.1) & 0.2 (0.0) & ---\\
\cmidrule(r){2-9}
 & clDice & baseline & 34.7 (5.2) & 19.6 (4.9) & --- & 0.7 (0.1) & 0.3 (0.0) & ---\\
 &  & +\textbf{CoLeTra} & \textbf{34.3} (5.4) & \textbf{19.1} (3.3) & --- & 0.7 (0.1) & \textbf{0.2} (0.0) & ---\\
\cmidrule(r){2-9}
 & Warploss & baseline & 75.5 (19.4) & 29.2 (3.4) & --- & 1.0 (0.2) & 0.2 (0.0) & ---\\
 &  & +\textbf{CoLeTra} & \textbf{70.2} (16.6) & \textbf{28.0} (3.0) & --- & \textbf{0.9} (0.2) & \textbf{0.2} (0.0) & ---\\
\cmidrule(r){2-9}
 & Topoloss & baseline & 121.0 (13.6) & 34.9 (3.1) & --- & 1.4 (0.1) & 0.2 (0.0) & ---\\
 &  & +\textbf{CoLeTra} & \textbf{120.0} (13.7) & \textbf{32.7} (2.5) & --- & \textbf{1.4} (0.2) & \textbf{0.2} (0.0) & ---\\
\midrule
\parbox[t]{2mm}{\multirow{12}{*}{\rotatebox[origin=c]{90}{\shortstack[c]{Crack500}}}} & CE & baseline & 12.28 (4.623) & 4.345 (2.351) & --- & 0.164 (0.032) & 0.043 (0.018) & ---\\
 &  & +\textbf{CoLeTra} & \textbf{11.375} (4.699) & \textbf{3.925} (1.842) & --- & \textbf{0.156} (0.031) & \textbf{0.039} (0.014) & ---\\
\cmidrule(r){2-9}
 & Dice & baseline & 10.278 (3.085) & 3.085 (1.047) & --- & 0.138 (0.021) & 0.029 (0.008) & ---\\
 &  & +\textbf{CoLeTra} & \textbf{10.082} (2.601) & \textbf{2.968} (1.077) & --- & \textbf{0.133} (0.017) & \textbf{0.028} (0.008) & ---\\
\cmidrule(r){2-9}
 & RWLoss & baseline & 10.852 (4.694) & 3.353 (1.547) & --- & 0.146 (0.03) & 0.031 (0.011) & ---\\
 &  & +\textbf{CoLeTra} & \textbf{9.853} (3.719) & \textbf{3.342} (1.416) & --- & \textbf{0.138} (0.023) & \textbf{0.03} (0.01) & ---\\
\cmidrule(r){2-9}
 & clDice & baseline & 10.885 (2.803) & 68.645 (85.574) & --- & 0.134 (0.023) & 0.473 (0.578) & ---\\
 &  & +\textbf{CoLeTra} & \textbf{10.377} (1.586) & 91.62 (87.311) & --- & \textbf{0.129} (0.016) & 0.627 (0.59) & ---\\
\cmidrule(r){2-9}
 & Warploss & baseline & 10.278 (3.085) & 3.085 (1.047) & --- & 0.138 (0.021) & 0.029 (0.008) & ---\\
 &  & +\textbf{CoLeTra} & \textbf{10.088} (2.692) & 3.108 (1.07) & --- & \textbf{0.134} (0.019) & \textbf{0.028} (0.007) & ---\\
\cmidrule(r){2-9}
 & Topoloss & baseline & 12.28 (4.623) & 4.345 (2.351) & --- & 0.164 (0.032) & 0.043 (0.018) & ---\\
 &  & +\textbf{CoLeTra} & \textbf{11.812} (4.855) & \textbf{4.325} (2.327) & --- & \textbf{0.162} (0.031) & 0.043 (0.018) & ---\\
\midrule
\parbox[t]{2mm}{\multirow{12}{*}{\rotatebox[origin=c]{90}{\shortstack[c]{PulmonaryVA}}}} & CE & baseline & 503 (54) & 159 (79) & 22 (15) & 23 (1) & 5 (2) & 2 (0)\\
 &  & +\textbf{CoLeTra} & \textbf{397} (74) & \textbf{68} (37) & \textbf{15} (10) & \textbf{21} (4) & \textbf{4} (1) & 3 (0)\\
\cmidrule(r){2-9}
 & Dice & baseline & 753 (309) & 345 (279) & 83 (98) & 26 (5) & 8 (5) & 3 (2)\\
 &  & +\textbf{CoLeTra} & \textbf{277} (107) & \textbf{33} (21) & \textbf{11} (3) & \textbf{17} (3) & \textbf{5} (0) & \textbf{3} (0)\\
\cmidrule(r){2-9}
 & RWLoss & baseline & 486 (215) & 108 (116) & 17 (16) & 18 (5) & 4 (1) & 2 (0)\\
 &  & +\textbf{CoLeTra} & 575 (371) & \textbf{100} (106) & \textbf{13} (6) & 20 (4) & 5 (1) & 3 (0)\\
\cmidrule(r){2-9}
 & clDice & baseline & 192 (59) & 29 (20) & 14 (6) & 8 (1) & 3 (0) & 3 (0)\\
 &  & +\textbf{CoLeTra} & \textbf{129} (31) & 34 (7) & \textbf{14} (1) & 9 (1) & 5 (0) & 3 (0)\\
\cmidrule(r){2-9}
 & Warploss & baseline & 549 (126) & 125 (85) & 17 (12) & 22 (4) & 5 (1) & 2 (0)\\
 &  & +\textbf{CoLeTra} & \textbf{277} (107) & \textbf{33} (21) & \textbf{11} (3) & \textbf{17} (3) & 5 (0) & 3 (0)\\
\cmidrule(r){2-9}
 & Topoloss & baseline & 522 (50) & 187 (49) & 25 (12) & 23 (2) & 6 (1) & 2 (0)\\
 &  & +\textbf{CoLeTra} & \textbf{378} (87) & \textbf{57} (39) & \textbf{11} (6) & \textbf{22} (3) & \textbf{3} (1) & 3 (0)\\
\midrule
\parbox[t]{2mm}{\multirow{12}{*}{\rotatebox[origin=c]{90}{\shortstack[c]{Narwhal}}}} & CE & baseline & 24430 (4492) & 888 (450) & 58 (39) & 151 (32) & 7 (1) & 0 (0)\\
 &  & +\textbf{CoLeTra} & \textbf{16806} (4805) & \textbf{810} (396) & 255 (359) & \textbf{102} (33) & \textbf{6} (1) & 1 (2)\\
\cmidrule(r){2-9}
 & Dice & baseline & 20969 (4201) & 736 (511) & 44 (17) & 124 (33) & 5 (1) & 0 (0)\\
 &  & +\textbf{CoLeTra} & 24830 (10180) & 765 (495) & 191 (241) & 146 (69) & 5 (1) & 1 (1)\\
\cmidrule(r){2-9}
 & RWLoss & baseline & 8112 (4885) & 718 (448) & 36 (13) & 44 (30) & 6 (1) & 0 (0)\\
 &  & +\textbf{CoLeTra} & \textbf{7286} (5773) & \textbf{673} (466) & \textbf{34} (14) & \textbf{43} (28) & \textbf{5} (1) & 0 (0)\\
\cmidrule(r){2-9}
 & clDice & baseline & 5221 (3044) & 620 (316) & 37 (9) & 22 (17) & 6 (0) & 0 (0)\\
 &  & +\textbf{CoLeTra} & \textbf{4159} (965) & \textbf{610} (335) & 39 (11) & \textbf{16} (4) & 6 (0) & 0 (0)\\
\cmidrule(r){2-9}
 & Warploss & baseline & 20966 (4222) & 735 (511) & 44 (17) & 124 (33) & 5 (1) & 0 (0)\\
 &  & +\textbf{CoLeTra} & 24844 (10194) & 765 (494) & 190 (240) & 146 (69) & 5 (1) & 1 (1)\\
\cmidrule(r){2-9}
 & Topoloss & baseline & 24431 (4494) & 888 (451) & 57 (39) & 151 (32) & 7 (1) & 0 (0)\\
 &  & +\textbf{CoLeTra} & \textbf{16779} (4860) & \textbf{822} (393) & 267 (379) & \textbf{102} (33) & \textbf{6} (1) & 1 (2)\\
\bottomrule
\end{tabular}
}
\end{table*}

\newpage \clearpage
\section{Performance metrics} \label{app:performance}
\Cref{tab:main_results} shows the global Betti 0 error ($\beta_{0e}$), Dice coefficient and clDice of our baseline models trained with extensive data augmentation and when they, additionally, used our CoLeTra data augmentation strategy.
\Cref{tab:fprprec}, additionally, shows the False Positive Rate and Precision.
\Cref{tab:interval} shows the confidence intervals (95\%) and \Cref{tab:effect} shows the effect size (Cohen's d).

\section{PulmonaryVA dataset 3D renders} \label{app:pulmonary}
\begin{figure*} 
  \centering
  \includegraphics[width=0.3\textwidth]{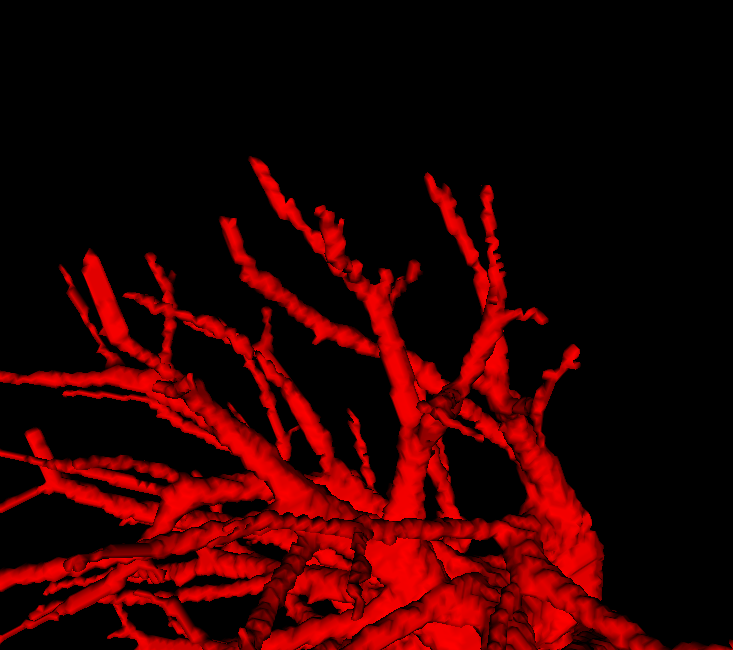}
  \includegraphics[width=0.3\textwidth]{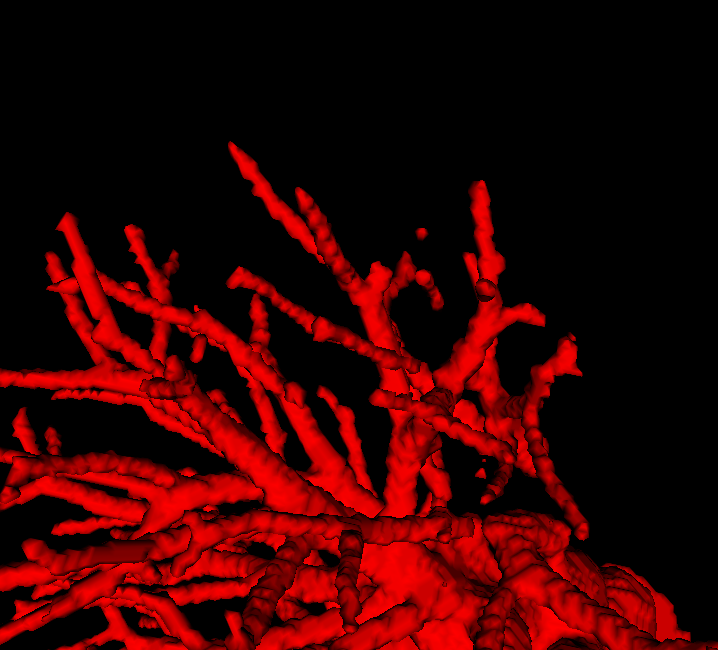}
  \includegraphics[width=0.3\textwidth]{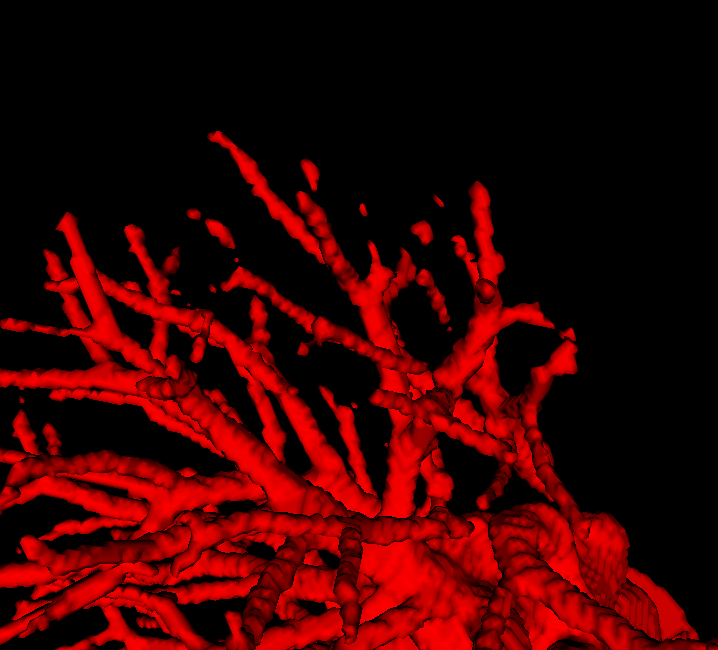}
  \caption{PumonaryVA dataset. Left: label. \textbf{Center: clDice + CoLeTra}. Right: clDice.} \label{fig:pulmonary_render}
\end{figure*}

\Cref{fig:pulmonary_render} illustrates a representative case where CoLeTra improves topology accuracy and connectivity in the blood vessels of the Pulmonary dataset.

\begin{table*}[h]
\caption{Performance comparison between baseline models (trained with extensive data augmentation) and our CoLeTra on the two architectures, three datasets, and six loss functions.}
\label{tab:main_results}
\centering
\tiny{
\begin{tabular}{lllllllll}
\toprule
\multicolumn{3}{c}{} & \multicolumn{3}{c}{DynUNet} & \multicolumn{3}{c}{AttentionUNet} \\
\cmidrule(r){4-6}
\cmidrule(r){7-9}
\multicolumn{3}{c}{} & $\beta_{0e}$ & Dice & clDice & $\beta_{0e}$ & Dice & clDice\\
\cmidrule(r){4-6}
\cmidrule(r){7-9}
\parbox[t]{2mm}{\multirow{12}{*}{\rotatebox[origin=c]{90}{\shortstack[c]{DRIVE}}}} & CE & baseline & 114.2 (11.8) & 72.6 (0.4) & 70.0 (0.7) & 121.0 (13.6) & 73.1 (0.5) & 70.7 (0.7)\\
 &  & +\textbf{CoLeTra} & \textbf{108.6} (10.2) & 72.5 (0.5) & \textbf{70.1} (0.6) & \textbf{118.9} (13.7) & 73.1 (0.5) & \textbf{70.9} (0.7)\\
\cmidrule(r){2-6}
\cmidrule(r){7-9}
 & Dice & baseline & 101.2 (10.8) & 73.7 (0.4) & 71.8 (0.5) & 75.5 (19.4) & 74.9 (0.6) & 73.4 (0.9)\\
 &  & +\textbf{CoLeTra} & \textbf{93.2} (9.2) & 73.7 (0.5) & \textbf{72.0} (0.6) & \textbf{69.4} (16.7) & 74.7 (0.5) & 73.4 (0.6)\\
\cmidrule(r){2-6}
\cmidrule(r){7-9}
 & RWLoss & baseline & 77.0 (8.2) & 74.4 (0.4) & 73.0 (0.6) & 58.7 (11.7) & 75.0 (0.3) & 73.8 (0.4)\\
 &  & +\textbf{CoLeTra} & \textbf{73.8} (6.2) & 74.4 (0.3) & 72.9 (0.5) & \textbf{56.8} (11.2) & 75.0 (0.5) & 73.8 (0.7)\\
\cmidrule(r){2-6}
\cmidrule(r){7-9}
 & clDice & baseline & 42.8 (7.3) & 73.8 (0.3) & 74.5 (0.4) & 34.7 (5.2) & 74.8 (0.3) & 76.5 (0.4)\\
 &  & +\textbf{CoLeTra} & \textbf{41.6} (6.6) & 73.5 (0.3) & \textbf{74.6} (0.4) & \textbf{34.3} (5.4) & 74.7 (0.3) & 76.4 (0.4)\\
\cmidrule(r){2-6}
\cmidrule(r){7-9}
 & Warploss & baseline & 99.2 (9.2) & 73.7 (0.4) & 71.8 (0.5) & 75.5 (19.4) & 74.9 (0.6) & 73.4 (0.9)\\
 &  & +\textbf{CoLeTra} & \textbf{94.5} (10.0) & 73.7 (0.5) & \textbf{72.0} (0.6) & \textbf{70.2} (16.6) & 74.7 (0.4) & 73.4 (0.5)\\
\cmidrule(r){2-6}
\cmidrule(r){7-9}
 & Topoloss & baseline & 111.2 (13.4) & 72.6 (0.5) & 70.0 (0.8) & 121.0 (13.6) & 73.1 (0.5) & 70.7 (0.7)\\
 &  & +\textbf{CoLeTra} & \textbf{110.4} (11.3) & 72.5 (0.5) & \textbf{70.1} (0.6) & \textbf{120.0} (13.7) & 73.1 (0.5) & \textbf{70.8} (0.7)\\
\midrule
\parbox[t]{2mm}{\multirow{12}{*}{\rotatebox[origin=c]{90}{\shortstack[c]{Crack500}}}} & CE & baseline & 9.687 (2.61) & 69.3 (1.4) & 78.2 (1.6) & 12.28 (4.62) & 67.4 (2.3) & 75.5 (2.5)\\
 &  & +\textbf{CoLeTra} & \textbf{9.135} (2.38) & \textbf{69.7} (1.4) & \textbf{78.6} (1.5) & \textbf{11.375} (4.7) & \textbf{68.5} (2.1) & \textbf{76.6} (2.3)\\
\cmidrule(r){2-6}
\cmidrule(r){7-9}
 & Dice & baseline & 9.422 (1.86) & 70.8 (1.0) & 79.7 (1.3) & 10.278 (3.08) & 69.3 (1.9) & 77.8 (2.0)\\
 &  & +\textbf{CoLeTra} & 9.458 (1.66) & \textbf{71.4} (1.0) & \textbf{80.1} (1.2) & \textbf{10.082} (2.6) & \textbf{70.2} (1.5) & \textbf{78.5} (1.7)\\
\cmidrule(r){2-6}
\cmidrule(r){7-9}
 & RWLoss & baseline & 9.313 (1.79) & 69.7 (1.1) & 78.6 (1.3) & 10.852 (4.69) & 68.3 (2.3) & 76.3 (2.4)\\
 &  & +\textbf{CoLeTra} & \textbf{9.177} (1.58) & \textbf{70.2} (1.3) & \textbf{79.1} (1.4) & \textbf{9.853} (3.72) & 68.2 (1.5) & \textbf{76.4} (1.8)\\
\cmidrule(r){2-6}
\cmidrule(r){7-9}
 & clDice & baseline & 10.337 (0.96) & 70.4 (1.1) & 82.6 (1.5) & 10.885 (2.8) & 67.9 (2.8) & 80.8 (2.8)\\
 &  & +\textbf{CoLeTra} & 10.522 (0.89) & \textbf{70.8} (1.0) & \textbf{83.1} (1.3) & \textbf{10.377} (1.59) & 67.1 (3.1) & \textbf{81.1} (2.7)\\
\cmidrule(r){2-6}
\cmidrule(r){7-9}
 & Warploss & baseline & 9.573 (1.76) & 70.8 (1.1) & 79.8 (1.3) & 10.278 (3.08) & 69.3 (1.9) & 77.8 (2.0)\\
 &  & +\textbf{CoLeTra} & \textbf{9.535} (1.69) & \textbf{71.3} (1.1) & \textbf{80.0} (1.3) & \textbf{10.088} (2.69) & \textbf{70.0} (1.5) & \textbf{78.3} (1.6)\\
\cmidrule(r){2-6}
\cmidrule(r){7-9}
 & Topoloss & baseline & 9.653 (2.35) & 69.3 (1.4) & 78.2 (1.6) & 12.28 (4.62) & 67.4 (2.3) & 75.5 (2.5)\\
 &  & +\textbf{CoLeTra} & \textbf{9.288} (2.36) & \textbf{69.7} (1.5) & \textbf{78.5} (1.6) & \textbf{11.812} (4.85) & \textbf{68.2} (2.1) & \textbf{76.1} (2.3)\\
\midrule
\parbox[t]{2mm}{\multirow{12}{*}{\rotatebox[origin=c]{90}{\shortstack[c]{PulmonaryVA}}}} & CE & baseline & 247 (14) & 76.0 (1.5) & 82.7 (1.1) & 503 (54) & 56.4 (2.2) & 65.0 (3.5)\\
 &  & +\textbf{CoLeTra} & \textbf{228} (15) & \textbf{77.8} (1.8) & \textbf{83.8} (1.2) & \textbf{397} (74) & 56.3 (3.7) & \textbf{68.7} (3.1)\\
\cmidrule(r){2-6}
\cmidrule(r){7-9}
 & Dice & baseline & 306 (27) & 75.0 (2.7) & 81.0 (2.1) & 753 (309) & 50.2 (11.3) & 55.1 (12.7)\\
 &  & +\textbf{CoLeTra} & \textbf{290} (25) & \textbf{77.1} (1.8) & \textbf{82.7} (1.0) & \textbf{277} (107) & \textbf{58.9} (3.7) & \textbf{73.9} (3.8)\\
\cmidrule(r){2-6}
\cmidrule(r){7-9}
 & RWLoss & baseline & 281 (26) & 71.9 (3.4) & 81.8 (1.9) & 486 (215) & 58.9 (5.4) & 66.2 (9.5)\\
 &  & +\textbf{CoLeTra} & \textbf{265} (21) & 71.2 (2.8) & 81.6 (1.0) & 575 (371) & 55.3 (4.0) & \textbf{68.2} (6.2)\\
\cmidrule(r){2-6}
\cmidrule(r){7-9}
 & clDice & baseline & 78 (12) & 76.4 (2.4) & 85.1 (1.9) & 192 (59) & 52.0 (7.3) & 76.4 (4.9)\\
 &  & +\textbf{CoLeTra} & \textbf{63} (6) & 76.1 (2.3) & \textbf{85.7} (1.5) & \textbf{129} (31) & \textbf{59.0} (2.0) & \textbf{80.2} (1.5)\\
\cmidrule(r){2-6}
\cmidrule(r){7-9}
 & Warploss & baseline & 305 (31) & 76.7 (2.8) & 81.7 (2.0) & 549 (126) & 58.3 (4.7) & 66.5 (6.9)\\
 &  & +\textbf{CoLeTra} & \textbf{289} (26) & \textbf{77.2} (1.9) & \textbf{82.8} (1.1) & \textbf{277} (107) & \textbf{58.9} (3.7) & \textbf{73.9} (3.8)\\
\cmidrule(r){2-6}
\cmidrule(r){7-9}
 & Topoloss & baseline & 245 (13) & 76.2 (1.5) & 82.7 (1.2) & 522 (50) & 57.4 (2.5) & 63.8 (2.6)\\
 &  & +\textbf{CoLeTra} & \textbf{227} (15) & \textbf{77.8} (1.8) & \textbf{83.8} (1.2) & \textbf{378} (87) & 57.4 (2.6) & \textbf{69.9} (3.9)\\
\midrule
\parbox[t]{2mm}{\multirow{12}{*}{\rotatebox[origin=c]{90}{\shortstack[c]{Narwhal}}}} & CE & baseline & 9907 (1128) & 67.1 (5.0) & 85.5 (1.3) & 24430 (4492) & 70.1 (3.1) & 83.4 (2.0)\\
 &  & +\textbf{CoLeTra} & \textbf{8988} (429) & \textbf{68.6} (4.4) & 85.1 (1.4) & \textbf{16806} (4805) & \textbf{72.6} (1.0) & \textbf{85.1} (0.7)\\
\cmidrule(r){2-6}
\cmidrule(r){7-9}
 & Dice & baseline & 20280 (249) & 31.4 (0.4) & 67.0 (1.9) & 20969 (4201) & 49.4 (16.7) & 77.5 (6.6)\\
 &  & +\textbf{CoLeTra} & \textbf{18457} (1220) & \textbf{31.8} (2.0) & 66.3 (2.8) & 24830 (10180) & 46.5 (14.6) & 73.8 (8.7)\\
\cmidrule(r){2-6}
\cmidrule(r){7-9}
 & RWLoss & baseline & 12697 (2712) & 63.2 (2.0) & 83.3 (0.3) & 8112 (4885) & 54.5 (18.2) & 83.7 (4.0)\\
 &  & +\textbf{CoLeTra} & 13046 (3102) & 62.1 (0.5) & 82.9 (0.2) & \textbf{7286} (5773) & 53.5 (19.1) & 83.7 (3.1)\\
\cmidrule(r){2-6}
\cmidrule(r){7-9}
 & clDice & baseline & 16328 (1693) & 34.4 (4.0) & 66.1 (4.4) & 5221 (3044) & 41.4 (9.4) & 78.7 (4.4)\\
 &  & +\textbf{CoLeTra} & \textbf{15743} (592) & \textbf{36.2} (2.1) & \textbf{67.4} (2.5) & \textbf{4159} (965) & 39.7 (5.5) & \textbf{80.4} (4.1)\\
\cmidrule(r){2-6}
\cmidrule(r){7-9}
 & Warploss & baseline & 18922 (1388) & 32.2 (1.5) & 67.5 (1.8) & 20966 (4222) & 49.5 (16.7) & 77.5 (6.6)\\
 &  & +\textbf{CoLeTra} & \textbf{18698} (781) & 31.3 (2.5) & 66.2 (2.9) & 24844 (10194) & 46.5 (14.6) & 73.8 (8.7)\\
\cmidrule(r){2-6}
\cmidrule(r){7-9}
 & Topoloss & baseline & 10127 (1244) & 66.9 (5.3) & 84.8 (2.2) & 24431 (4494) & 70.1 (3.1) & 83.4 (2.0)\\
 &  & +\textbf{CoLeTra} & \textbf{9041} (682) & \textbf{67.1} (4.6) & 84.6 (1.0) & \textbf{16779} (4860) & \textbf{72.6} (1.0) & \textbf{85.1} (0.7)\\
\bottomrule
\end{tabular}
}
\end{table*}

\begin{table*}
\caption{False positive rate (FPR) and Precision.}
\label{tab:fprprec}
\centering
\tiny{
\begin{tabular}{lllllll}
\toprule
\multicolumn{3}{c}{} & \multicolumn{2}{c}{DynUNet} & \multicolumn{2}{c}{AttentionUNet} \\
\cmidrule(r){4-5}
\cmidrule(r){6-7}
\multicolumn{3}{c}{} & FPR & Precision & FPR & Precision\\
\cmidrule(r){4-5}
\cmidrule(r){6-7}
\parbox[t]{2mm}{\multirow{12}{*}{\rotatebox[origin=c]{90}{\shortstack[c]{DRIVE}}}} & CE & baseline & 2.1 (0.1) & 76.0 (0.8) & 2.1 (0.1) & 76.1 (0.8)\\
 &  & +\textbf{CoLeTra} & 2.1 (0.1) & 75.8 (0.9) & 2.2 (0.1) & 75.2 (1.0)\\
\cmidrule(r){2-5}
\cmidrule(r){6-7}
 & Dice & baseline & 2.3 (0.1) & 75.1 (0.6) & 2.2 (0.1) & 76.0 (0.7)\\
 &  & +\textbf{CoLeTra} & 2.4 (0.1) & 74.2 (0.7) & 2.4 (0.1) & 75.0 (0.9)\\
\cmidrule(r){2-5}
\cmidrule(r){6-7}
 & RWLoss & baseline & 2.2 (0.1) & 76.0 (0.9) & 2.1 (0.1) & 76.6 (0.8)\\
 &  & +\textbf{CoLeTra} & 2.3 (0.1) & 75.4 (0.7) & 2.2 (0.2) & 76.3 (1.0)\\
\cmidrule(r){2-5}
\cmidrule(r){6-7}
 & clDice & baseline & 2.9 (0.1) & 71.4 (0.5) & 3.0 (0.1) & 71.4 (0.7)\\
 &  & +\textbf{CoLeTra} & 3.1 (0.1) & 70.4 (0.6) & 3.1 (0.1) & 70.8 (0.5)\\
\cmidrule(r){2-5}
\cmidrule(r){6-7}
 & Warploss & baseline & 2.3 (0.1) & 75.1 (0.6) & 2.2 (0.1) & 76.0 (0.7)\\
 &  & +\textbf{CoLeTra} & 2.4 (0.1) & 74.1 (0.7) & 2.4 (0.1) & 74.9 (0.8)\\
\cmidrule(r){2-5}
\cmidrule(r){6-7}
 & Topoloss & baseline & 2.1 (0.1) & 76.1 (0.8) & 2.1 (0.1) & 76.1 (0.8)\\
 &  & +\textbf{CoLeTra} & 2.1 (0.1) & 75.9 (0.9) & 2.2 (0.2) & 75.2 (1.0)\\
\midrule
\parbox[t]{2mm}{\multirow{12}{*}{\rotatebox[origin=c]{90}{\shortstack[c]{Crack500}}}} & CE & baseline & 0.6 (0.1) & 74.7 (1.6) & 0.6 (0.1) & 74.7 (2.2)\\
 &  & +\textbf{CoLeTra} & 0.6 (0.1) & 74.5 (1.6) & 0.6 (0.1) & 74.5 (2.2)\\
\cmidrule(r){2-5}
\cmidrule(r){6-7}
 & Dice & baseline & 0.8 (0.1) & 72.2 (1.3) & 0.8 (0.1) & 72.0 (2.1)\\
 &  & +\textbf{CoLeTra} & 0.8 (0.1) & 71.4 (1.3) & 0.8 (0.1) & 71.4 (1.9)\\
\cmidrule(r){2-5}
\cmidrule(r){6-7}
 & RWLoss & baseline & 0.7 (0.1) & 73.4 (1.4) & 0.7 (0.1) & 73.5 (2.2)\\
 &  & +\textbf{CoLeTra} & 0.8 (0.1) & 72.7 (1.5) & 0.7 (0.1) & 73.3 (1.5)\\
\cmidrule(r){2-5}
\cmidrule(r){6-7}
 & clDice & baseline & 1.1 (0.1) & 67.5 (1.3) & 1.3 (0.2) & 63.5 (4.0)\\
 &  & +\textbf{CoLeTra} & 1.1 (0.1) & 66.9 (1.2) & 1.4 (0.3) & 61.1 (5.5)\\
\cmidrule(r){2-5}
\cmidrule(r){6-7}
 & Warploss & baseline & 0.8 (0.1) & 72.2 (1.3) & 0.8 (0.1) & 72.0 (2.1)\\
 &  & +\textbf{CoLeTra} & 0.8 (0.1) & 71.3 (1.4) & 0.8 (0.1) & 71.6 (1.8)\\
\cmidrule(r){2-5}
\cmidrule(r){6-7}
 & Topoloss & baseline & 0.6 (0.1) & 74.7 (1.6) & 0.6 (0.1) & 74.7 (2.2)\\
 &  & +\textbf{CoLeTra} & 0.6 (0.1) & 74.6 (1.7) & 0.6 (0.1) & \textbf{74.8} (2.0)\\
\midrule
\parbox[t]{2mm}{\multirow{12}{*}{\rotatebox[origin=c]{90}{\shortstack[c]{PulmonaryVA}}}} & CE & baseline & 0.4 (0.1) & 71.7 (2.5) & 0.8 (0.1) & 47.7 (3.0)\\
 &  & +\textbf{CoLeTra} & \textbf{0.3} (0.1) & \textbf{76.7} (3.2) & \textbf{0.5} (0.2) & \textbf{58.1} (5.6)\\
\cmidrule(r){2-5}
\cmidrule(r){6-7}
 & Dice & baseline & 0.5 (0.1) & 66.9 (4.4) & 1.7 (0.9) & 38.1 (12.7)\\
 &  & +\textbf{CoLeTra} & \textbf{0.4} (0.1) & \textbf{71.2} (2.7) & \textbf{0.3} (0.3) & \textbf{70.4} (14.6)\\
\cmidrule(r){2-5}
\cmidrule(r){6-7}
 & RWLoss & baseline & 0.5 (0.1) & 65.5 (4.9) & 0.8 (0.4) & 51.8 (10.4)\\
 &  & +\textbf{CoLeTra} & \textbf{0.4} (0.1) & \textbf{66.7} (4.0) & \textbf{0.6} (0.3) & \textbf{57.0} (9.2)\\
\cmidrule(r){2-5}
\cmidrule(r){6-7}
 & clDice & baseline & 0.5 (0.1) & 67.9 (3.5) & 1.6 (0.8) & 39.5 (8.7)\\
 &  & +\textbf{CoLeTra} & 0.5 (0.1) & \textbf{68.4} (3.4) & \textbf{0.5} (0.1) & \textbf{57.4} (3.0)\\
\cmidrule(r){2-5}
\cmidrule(r){6-7}
 & Warploss & baseline & 0.4 (0.1) & 69.8 (4.7) & 1.0 (0.3) & 47.2 (7.0)\\
 &  & +\textbf{CoLeTra} & 0.4 (0.1) & \textbf{71.3} (2.9) & \textbf{0.3} (0.3) & \textbf{70.4} (14.6)\\
\cmidrule(r){2-5}
\cmidrule(r){6-7}
 & Topoloss & baseline & 0.4 (0.1) & 72.0 (2.8) & 0.8 (0.1) & 48.8 (3.9)\\
 &  & +\textbf{CoLeTra} & \textbf{0.3} (0.1) & \textbf{76.7} (3.2) & \textbf{0.5} (0.2) & \textbf{61.0} (6.7)\\
\midrule
\parbox[t]{2mm}{\multirow{12}{*}{\rotatebox[origin=c]{90}{\shortstack[c]{Narwhal}}}} & CE & baseline & 0.5 (0.1) & 80.6 (1.8) & 0.8 (0.3) & 75.1 (4.8)\\
 &  & +\textbf{CoLeTra} & 0.5 (0.2) & 79.4 (3.8) & \textbf{0.7} (0.2) & \textbf{76.5} (2.7)\\
\cmidrule(r){2-5}
\cmidrule(r){6-7}
 & Dice & baseline & 0.1 (0.0) & 85.6 (1.6) & 0.8 (1.2) & 76.7 (17.7)\\
 &  & +\textbf{CoLeTra} & 0.1 (0.0) & \textbf{86.1} (3.0) & 1.2 (1.8) & 73.0 (22.0)\\
\cmidrule(r){2-5}
\cmidrule(r){6-7}
 & RWLoss & baseline & 0.4 (0.1) & 81.9 (2.9) & 3.8 (5.6) & 59.3 (32.7)\\
 &  & +\textbf{CoLeTra} & 0.4 (0.0) & 79.6 (1.4) & 4.2 (6.4) & 58.7 (33.7)\\
\cmidrule(r){2-5}
\cmidrule(r){6-7}
 & clDice & baseline & 0.1 (0.0) & 86.5 (1.4) & 5.9 (3.3) & 31.8 (13.7)\\
 &  & +\textbf{CoLeTra} & 0.1 (0.0) & \textbf{87.1} (0.9) & 6.4 (2.2) & 28.1 (6.3)\\
\cmidrule(r){2-5}
\cmidrule(r){6-7}
 & Warploss & baseline & 0.1 (0.0) & 86.1 (2.6) & 0.8 (1.2) & 76.8 (17.7)\\
 &  & +\textbf{CoLeTra} & 0.1 (0.0) & 85.9 (3.0) & 1.2 (1.8) & 73.0 (22.0)\\
\cmidrule(r){2-5}
\cmidrule(r){6-7}
 & Topoloss & baseline & 0.5 (0.1) & 79.3 (2.1) & 0.8 (0.3) & 75.1 (4.8)\\
 &  & +\textbf{CoLeTra} & 0.5 (0.2) & \textbf{79.6} (3.4) & \textbf{0.7} (0.2) & \textbf{76.5} (2.7)\\
\bottomrule
\end{tabular}
}
\end{table*}

\begin{table*}
\caption{Confidence intervals (95\%)}
\label{tab:interval}
\centering
\tiny{
\begin{tabular}{lllllllll}
\toprule
\multicolumn{3}{c}{} & \multicolumn{3}{c}{DynUNet} & \multicolumn{3}{c}{AttentionUNet} \\
\cmidrule(r){4-6}
\cmidrule(r){7-9}
\multicolumn{3}{c}{} & $\beta_{0e}$ & Dice & clDice & Betti & Dice & clDice\\
\cmidrule(r){4-6}
\cmidrule(r){7-9}
\parbox[t]{2mm}{\multirow{12}{*}{\rotatebox[origin=c]{90}{\shortstack[c]{DRIVE}}}} & CE & baseline & [107.6,120.7] & [71.6,73.6] & [68.8,71.3] & [112.8,129.2] & [72.2,73.9] & [69.5,71.9]\\
 &  & +\textbf{CoLeTra} & [103.2,114.0] & [71.5,73.5] & [68.9,71.2] & [112.2,125.6] & [72.2,74.0] & [69.7,72.0]\\
\cmidrule(r){2-6}
\cmidrule(r){7-9}
 & Dice & baseline & [93.3,109.0] & [72.7,74.6] & [70.5,73.0] & [68.1,83.0] & [74.1,75.6] & [72.2,74.6]\\
 &  & +\textbf{CoLeTra} & [86.1,100.3] & [72.7,74.7] & [70.7,73.2] & [62.9,76.0] & [73.8,75.6] & [72.2,74.6]\\
\cmidrule(r){2-6}
\cmidrule(r){7-9}
 & RWLoss & baseline & [70.9,83.2] & [73.5,75.4] & [71.7,74.2] & [53.0,64.4] & [74.2,75.8] & [72.6,74.9]\\
 &  & +\textbf{CoLeTra} & [68.2,79.4] & [73.4,75.3] & [71.7,74.2] & [51.6,62.1] & [74.2,75.8] & [72.6,75.0]\\
\cmidrule(r){2-6}
\cmidrule(r){7-9}
 & clDice & baseline & [39.1,46.5] & [72.9,74.6] & [73.4,75.7] & [30.5,38.9] & [74.0,75.6] & [75.4,77.6]\\
 &  & +\textbf{CoLeTra} & [37.5,45.7] & [72.7,74.4] & [73.4,75.7] & [30.4,38.2] & [73.8,75.5] & [75.2,77.5]\\
\cmidrule(r){2-6}
\cmidrule(r){7-9}
 & Warploss & baseline & [91.1,107.3] & [72.7,74.6] & [70.6,73.1] & [68.1,83.0] & [74.1,75.6] & [72.2,74.6]\\
 &  & +\textbf{CoLeTra} & [87.4,101.7] & [72.7,74.7] & [70.7,73.2] & [63.9,76.4] & [73.8,75.6] & [72.2,74.6]\\
\cmidrule(r){2-6}
\cmidrule(r){7-9}
 & Topoloss & baseline & [104.5,117.9] & [71.6,73.6] & [68.8,71.2] & [112.8,129.2] & [72.2,73.9] & [69.5,71.9]\\
 &  & +\textbf{CoLeTra} & [104.4,116.5] & [71.5,73.5] & [68.9,71.3] & [112.7,127.4] & [72.2,74.0] & [69.7,72.0]\\
\midrule
\parbox[t]{2mm}{\multirow{12}{*}{\rotatebox[origin=c]{90}{\shortstack[c]{Crack500}}}} & CE & baseline & [8.146,11.227] & [68.4,70.3] & [77.2,79.3] & [10.902,13.658] & [66.3,68.5] & [74.3,76.7]\\
 &  & +\textbf{CoLeTra} & [7.707,10.563] & [68.8,70.6] & [77.5,79.6] & [10.094,12.656] & [67.4,69.5] & [75.4,77.8]\\
\cmidrule(r){2-6}
\cmidrule(r){7-9}
 & Dice & baseline & [7.769,11.075] & [69.9,71.7] & [78.7,80.7] & [8.673,11.884] & [68.3,70.3] & [76.7,78.9]\\
 &  & +\textbf{CoLeTra} & [7.749,11.168] & [70.6,72.3] & [79.1,81.1] & [8.445,11.718] & [69.2,71.1] & [77.4,79.6]\\
\cmidrule(r){2-6}
\cmidrule(r){7-9}
 & RWLoss & baseline & [7.635,10.991] & [68.7,70.6] & [77.5,79.7] & [8.938,12.765] & [67.2,69.3] & [75.1,77.5]\\
 &  & +\textbf{CoLeTra} & [7.451,10.902] & [69.3,71.1] & [78.0,80.1] & [8.358,11.348] & [67.1,69.3] & [75.2,77.6]\\
\cmidrule(r){2-6}
\cmidrule(r){7-9}
 & clDice & baseline & [8.394,12.28] & [69.6,71.3] & [81.7,83.5] & [8.647,13.123] & [66.9,68.9] & [79.8,81.8]\\
 &  & +\textbf{CoLeTra} & [8.587,12.456] & [69.9,71.6] & [82.2,84.0] & [8.458,12.295] & [66.1,68.1] & [80.1,82.1]\\
\cmidrule(r){2-6}
\cmidrule(r){7-9}
 & Warploss & baseline & [7.922,11.225] & [69.9,71.7] & [78.8,80.7] & [8.673,11.884] & [68.3,70.3] & [76.7,78.9]\\
 &  & +\textbf{CoLeTra} & [7.789,11.281] & [70.4,72.2] & [79.0,81.0] & [8.46,11.716] & [69.0,70.9] & [77.2,79.4]\\
\cmidrule(r){2-6}
\cmidrule(r){7-9}
 & Topoloss & baseline & [8.113,11.194] & [68.4,70.2] & [77.2,79.3] & [10.902,13.658] & [66.3,68.5] & [74.3,76.7]\\
 &  & +\textbf{CoLeTra} & [7.83,10.746] & [68.8,70.6] & [77.5,79.6] & [10.463,13.16] & [67.1,69.3] & [74.9,77.3]\\
\midrule
\parbox[t]{2mm}{\multirow{12}{*}{\rotatebox[origin=c]{90}{\shortstack[c]{PulmonaryVA}}}} & CE & baseline & [224,270] & [74.1,77.9] & [81.4,83.9] & [460,548] & [54.8,57.9] & [63.4,66.5]\\
 &  & +\textbf{CoLeTra} & [207,249] & [76.3,79.3] & [82.7,84.8] & [359,435] & [54.1,58.5] & [66.7,70.7]\\
\cmidrule(r){2-6}
\cmidrule(r){7-9}
 & Dice & baseline & [277,337] & [72.8,77.2] & [79.4,82.6] & [615,891] & [47.3,53.1] & [51.7,58.5]\\
 &  & +\textbf{CoLeTra} & [265,316] & [75.2,79.0] & [81.4,84.0] & [237,318] & [57.0,60.8] & [72.2,75.5]\\
\cmidrule(r){2-6}
\cmidrule(r){7-9}
 & RWLoss & baseline & [256,307] & [70.0,73.7] & [80.7,82.9] & [418,555] & [57.2,60.6] & [63.8,68.6]\\
 &  & +\textbf{CoLeTra} & [243,289] & [69.3,73.0] & [80.5,82.8] & [382,768] & [52.6,57.9] & [64.9,71.5]\\
\cmidrule(r){2-6}
\cmidrule(r){7-9}
 & clDice & baseline & [66,90] & [74.8,78.0] & [83.9,86.3] & [153,233] & [49.8,54.2] & [73.7,79.1]\\
 &  & +\textbf{CoLeTra} & [53,73] & [74.3,77.9] & [84.6,86.8] & [97,163] & [56.9,61.1] & [78.8,81.5]\\
\cmidrule(r){2-6}
\cmidrule(r){7-9}
 & Warploss & baseline & [276,336] & [74.8,78.6] & [80.2,83.2] & [473,626] & [56.5,60.1] & [64.1,68.8]\\
 &  & +\textbf{CoLeTra} & [265,314] & [75.3,79.1] & [81.5,84.0] & [237,318] & [57.0,60.8] & [72.2,75.5]\\
\cmidrule(r){2-6}
\cmidrule(r){7-9}
 & Topoloss & baseline & [221,269] & [74.4,78.0] & [81.5,83.9] & [478,567] & [55.8,59.0] & [62.2,65.4]\\
 &  & +\textbf{CoLeTra} & [207,249] & [76.3,79.3] & [82.8,84.8] & [342,416] & [55.2,59.6] & [67.9,72.0]\\
\midrule
\parbox[t]{2mm}{\multirow{12}{*}{\rotatebox[origin=c]{90}{\shortstack[c]{Narwhal}}}} & CE & baseline & [8570,11246] & [63.6,70.7] & [84.1,86.8] & [20684,28178] & [67.4,72.7] & [81.6,85.2]\\
 &  & +\textbf{CoLeTra} & [7994,9983] & [65.4,71.7] & [83.7,86.6] & [13023,20589] & [71.0,74.2] & [83.9,86.3]\\
\cmidrule(r){2-6}
\cmidrule(r){7-9}
 & Dice & baseline & [19758,20802] & [30.1,32.6] & [64.1,70.0] & [17864,24075] & [38.2,60.7] & [72.7,82.3]\\
 &  & +\textbf{CoLeTra} & [17539,19375] & [30.0,33.6] & [63.2,69.5] & [17916,31744] & [36.6,56.3] & [67.6,79.9]\\
\cmidrule(r){2-6}
\cmidrule(r){7-9}
 & RWLoss & baseline & [10633,14761] & [61.3,65.1] & [82.1,84.5] & [4739,11486] & [42.3,66.8] & [80.7,86.8]\\
 &  & +\textbf{CoLeTra} & [10776,15317] & [60.6,63.5] & [81.6,84.1] & [3349,11224] & [40.6,66.3] & [81.3,86.2]\\
\cmidrule(r){2-6}
\cmidrule(r){7-9}
 & clDice & baseline & [15083,17574] & [31.5,37.3] & [62.4,69.9] & [3139,7303] & [35.0,47.9] & [75.1,82.3]\\
 &  & +\textbf{CoLeTra} & [15124,16364] & [34.3,38.1] & [64.6,70.3] & [3354,4964] & [35.8,43.6] & [77.1,83.7]\\
\cmidrule(r){2-6}
\cmidrule(r){7-9}
 & Warploss & baseline & [17874,19971] & [30.7,33.8] & [64.6,70.4] & [17847,24085] & [38.2,60.7] & [72.7,82.3]\\
 &  & +\textbf{CoLeTra} & [17982,19415] & [29.3,33.3] & [63.0,69.4] & [17921,31767] & [36.6,56.3] & [67.6,79.9]\\
\cmidrule(r){2-6}
\cmidrule(r){7-9}
 & Topoloss & baseline & [8658,11597] & [63.2,70.7] & [83.0,86.6] & [20683,28179] & [67.4,72.7] & [81.6,85.2]\\
 &  & +\textbf{CoLeTra} & [8054,10028] & [63.7,70.4] & [83.4,85.9] & [12967,20591] & [71.0,74.2] & [83.9,86.3]\\
\bottomrule
\end{tabular}
}
\end{table*}

\begin{table*}
\caption{Effect size (Cohen's d).}
\label{tab:effect}
\centering
\tiny{
\begin{tabular}{lllllllll}
\toprule
\multicolumn{3}{c}{} & \multicolumn{3}{c}{DynUNet} & \multicolumn{3}{c}{AttentionUNet} \\
\cmidrule(r){4-6}
\cmidrule(r){7-9}
\multicolumn{3}{c}{} & Betti & Dice & clDice & Betti & Dice & clDice\\
\cmidrule(r){4-6}
\cmidrule(r){7-9}
\parbox[t]{2mm}{\multirow{12}{*}{\rotatebox[origin=c]{90}{\shortstack[c]{DRIVE}}}} & CE & baseline & \multirow{2}{*}{0.239} & \multirow{2}{*}{0.022} & \multirow{2}{*}{-0.005} & \multirow{2}{*}{0.074} & \multirow{2}{*}{-0.024} & \multirow{2}{*}{-0.035}\\
&  & +\textbf{CoLeTra} & & & & & & \\
\cmidrule(r){2-6}
\cmidrule(r){7-9}
& Dice & baseline & \multirow{2}{*}{0.277} & \multirow{2}{*}{-0.008} & \multirow{2}{*}{-0.047} & \multirow{2}{*}{0.226} & \multirow{2}{*}{0.038} & \multirow{2}{*}{-0.014}\\
&  & +\textbf{CoLeTra} & & & & & & \\
\cmidrule(r){2-6}
\cmidrule(r){7-9}
& RWLoss & baseline & \multirow{2}{*}{0.144} & \multirow{2}{*}{0.004} & \multirow{2}{*}{0.002} & \multirow{2}{*}{0.09} & \multirow{2}{*}{0.001} & \multirow{2}{*}{-0.015}\\
&  & +\textbf{CoLeTra} & & & & & & \\
\cmidrule(r){2-6}
\cmidrule(r){7-9}
& clDice & baseline & \multirow{2}{*}{0.079} & \multirow{2}{*}{0.076} & \multirow{2}{*}{-0.013} & \multirow{2}{*}{0.024} & \multirow{2}{*}{0.045} & \multirow{2}{*}{0.02}\\
&  & +\textbf{CoLeTra} & & & & & & \\
\cmidrule(r){2-6}
\cmidrule(r){7-9}
& Warploss & baseline & \multirow{2}{*}{0.16} & \multirow{2}{*}{0.002} & \multirow{2}{*}{-0.03} & \multirow{2}{*}{0.203} & \multirow{2}{*}{0.049} & \multirow{2}{*}{-0.014}\\
&  & +\textbf{CoLeTra} & & & & & & \\
\cmidrule(r){2-6}
\cmidrule(r){7-9}
& Topoloss & baseline & \multirow{2}{*}{0.031} & \multirow{2}{*}{0.017} & \multirow{2}{*}{-0.014} & \multirow{2}{*}{0.032} & \multirow{2}{*}{-0.023} & \multirow{2}{*}{-0.027}\\
&  & +\textbf{CoLeTra} & & & & & & \\
\midrule
\parbox[t]{2mm}{\multirow{12}{*}{\rotatebox[origin=c]{90}{\shortstack[c]{Crack500}}}} & CE & baseline & \multirow{2}{*}{0.03} & \multirow{2}{*}{-0.032} & \multirow{2}{*}{-0.028} & \multirow{2}{*}{0.055} & \multirow{2}{*}{-0.085} & \multirow{2}{*}{-0.074}\\
&  & +\textbf{CoLeTra} & & & & & & \\
\cmidrule(r){2-6}
\cmidrule(r){7-9}
& Dice & baseline & \multirow{2}{*}{-0.002} & \multirow{2}{*}{-0.06} & \multirow{2}{*}{-0.032} & \multirow{2}{*}{0.01} & \multirow{2}{*}{-0.069} & \multirow{2}{*}{-0.053}\\
&  & +\textbf{CoLeTra} & & & & & & \\
\cmidrule(r){2-6}
\cmidrule(r){7-9}
& RWLoss & baseline & \multirow{2}{*}{0.006} & \multirow{2}{*}{-0.045} & \multirow{2}{*}{-0.036} & \multirow{2}{*}{0.047} & \multirow{2}{*}{0.004} & \multirow{2}{*}{-0.005}\\
&  & +\textbf{CoLeTra} & & & & & & \\
\cmidrule(r){2-6}
\cmidrule(r){7-9}
& clDice & baseline & \multirow{2}{*}{-0.008} & \multirow{2}{*}{-0.032} & \multirow{2}{*}{-0.045} & \multirow{2}{*}{0.02} & \multirow{2}{*}{0.06} & \multirow{2}{*}{-0.021}\\
&  & +\textbf{CoLeTra} & & & & & & \\
\cmidrule(r){2-6}
\cmidrule(r){7-9}
& Warploss & baseline & \multirow{2}{*}{0.002} & \multirow{2}{*}{-0.045} & \multirow{2}{*}{-0.022} & \multirow{2}{*}{0.009} & \multirow{2}{*}{-0.052} & \multirow{2}{*}{-0.036}\\
&  & +\textbf{CoLeTra} & & & & & & \\
\cmidrule(r){2-6}
\cmidrule(r){7-9}
& Topoloss & baseline & \multirow{2}{*}{0.02} & \multirow{2}{*}{-0.032} & \multirow{2}{*}{-0.023} & \multirow{2}{*}{0.028} & \multirow{2}{*}{-0.061} & \multirow{2}{*}{-0.041}\\
&  & +\textbf{CoLeTra} & & & & & & \\
\midrule
\parbox[t]{2mm}{\multirow{12}{*}{\rotatebox[origin=c]{90}{\shortstack[c]{PulmonaryVA}}}} & CE & baseline & \multirow{2}{*}{0.21} & \multirow{2}{*}{-0.267} & \multirow{2}{*}{-0.244} & \multirow{2}{*}{0.645} & \multirow{2}{*}{0.005} & \multirow{2}{*}{-0.516}\\
&  & +\textbf{CoLeTra} & & & & & & \\
\cmidrule(r){2-6}
\cmidrule(r){7-9}
& Dice & baseline & \multirow{2}{*}{0.145} & \multirow{2}{*}{-0.259} & \multirow{2}{*}{-0.297} & \multirow{2}{*}{1.16} & \multirow{2}{*}{-0.873} & \multirow{2}{*}{-1.745}\\
&  & +\textbf{CoLeTra} & & & & & & \\
\cmidrule(r){2-6}
\cmidrule(r){7-9}
& RWLoss & baseline & \multirow{2}{*}{0.159} & \multirow{2}{*}{0.095} & \multirow{2}{*}{0.029} & \multirow{2}{*}{-0.152} & \multirow{2}{*}{0.409} & \multirow{2}{*}{-0.174}\\
&  & +\textbf{CoLeTra} & & & & & & \\
\cmidrule(r){2-6}
\cmidrule(r){7-9}
& clDice & baseline & \multirow{2}{*}{0.336} & \multirow{2}{*}{0.044} & \multirow{2}{*}{-0.131} & \multirow{2}{*}{0.426} & \multirow{2}{*}{-0.8} & \multirow{2}{*}{-0.439}\\
&  & +\textbf{CoLeTra} & & & & & & \\
\cmidrule(r){2-6}
\cmidrule(r){7-9}
& Warploss & baseline & \multirow{2}{*}{0.145} & \multirow{2}{*}{-0.063} & \multirow{2}{*}{-0.197} & \multirow{2}{*}{1.104} & \multirow{2}{*}{-0.075} & \multirow{2}{*}{-0.902}\\
&  & +\textbf{CoLeTra} & & & & & & \\
\cmidrule(r){2-6}
\cmidrule(r){7-9}
& Topoloss & baseline & \multirow{2}{*}{0.191} & \multirow{2}{*}{-0.241} & \multirow{2}{*}{-0.237} & \multirow{2}{*}{0.87} & \multirow{2}{*}{0.0} & \multirow{2}{*}{-0.833}\\
&  & +\textbf{CoLeTra} & & & & & & \\
\midrule
\parbox[t]{2mm}{\multirow{12}{*}{\rotatebox[origin=c]{90}{\shortstack[c]{Narwhal}}}} & CE & baseline & \multirow{2}{*}{0.636} & \multirow{2}{*}{-0.342} & \multirow{2}{*}{0.217} & \multirow{2}{*}{1.651} & \multirow{2}{*}{-0.931} & \multirow{2}{*}{-0.923}\\
&  & +\textbf{CoLeTra} & & & & & & \\
\cmidrule(r){2-6}
\cmidrule(r){7-9}
& Dice & baseline & \multirow{2}{*}{1.99} & \multirow{2}{*}{-0.246} & \multirow{2}{*}{0.187} & \multirow{2}{*}{-0.587} & \multirow{2}{*}{0.229} & \multirow{2}{*}{0.555}\\
&  & +\textbf{CoLeTra} & & & & & & \\
\cmidrule(r){2-6}
\cmidrule(r){7-9}
& RWLoss & baseline & \multirow{2}{*}{-0.131} & \multirow{2}{*}{0.554} & \multirow{2}{*}{0.278} & \multirow{2}{*}{0.184} & \multirow{2}{*}{0.07} & \multirow{2}{*}{0.001}\\
&  & +\textbf{CoLeTra} & & & & & & \\
\cmidrule(r){2-6}
\cmidrule(r){7-9}
& clDice & baseline & \multirow{2}{*}{0.484} & \multirow{2}{*}{-0.605} & \multirow{2}{*}{-0.322} & \multirow{2}{*}{0.549} & \multirow{2}{*}{0.269} & \multirow{2}{*}{-0.405}\\
&  & +\textbf{CoLeTra} & & & & & & \\
\cmidrule(r){2-6}
\cmidrule(r){7-9}
& Warploss & baseline & \multirow{2}{*}{0.203} & \multirow{2}{*}{0.425} & \multirow{2}{*}{0.353} & \multirow{2}{*}{-0.589} & \multirow{2}{*}{0.23} & \multirow{2}{*}{0.554}\\
&  & +\textbf{CoLeTra} & & & & & & \\
\cmidrule(r){2-6}
\cmidrule(r){7-9}
& Topoloss & baseline & \multirow{2}{*}{0.707} & \multirow{2}{*}{-0.036} & \multirow{2}{*}{0.096} & \multirow{2}{*}{1.65} & \multirow{2}{*}{-0.93} & \multirow{2}{*}{-0.92}\\
&  & +\textbf{CoLeTra} & & & & & & \\
\bottomrule
\end{tabular}
}
\end{table*}

\end{document}